\documentclass[sigconf]{acmart}

\AtBeginDocument{%
  \providecommand\BibTeX{{%
    \normalfont B\kern-0.5em{\scshape i\kern-0.25em b}\kern-0.8em\TeX}}}

\copyrightyear{2023}
\acmYear{2023}
\setcopyright{acmlicensed}\acmConference[MM '23]{Proceedings of the 31st ACM International Conference on Multimedia}{October 29-November 3, 2023}{Ottawa, ON, Canada}
\acmBooktitle{Proceedings of the 31st ACM International Conference on Multimedia (MM '23), October 29-November 3, 2023, Ottawa, ON, Canada}
\acmPrice{15.00}
\acmDOI{10.1145/3581783.3611809}
\acmISBN{979-8-4007-0108-5/23/10}

\usepackage{algorithm}
\usepackage{algorithmic}

\usepackage{subfig}
\usepackage{mathrsfs}

\usepackage{amssymb}
\usepackage{amsmath}
\usepackage{dsfont}
\usepackage{booktabs}
\usepackage{epstopdf}
\usepackage{multirow}
\usepackage{comment}
\usepackage{endnotes}
\usepackage{hyperref}
\usepackage{bibentry}
\usepackage{amsmath}
\usepackage{dsfont}
\usepackage{amsfonts}
\usepackage{amssymb}
\usepackage{bbm}
\usepackage{bbding}
\usepackage{makecell}
\usepackage[normalem]{ulem}
\useunder{\uline}{\ul}{}

\begin{document}
\title{CONVERT:~Contrastive Graph Clustering with Reliable Augmentation}
\author{Xihong Yang}
\email{xihong_edu@163.com}
\affiliation{%
  \institution{National University of Defense Technology}
  \city{Changsha}
  \state{Hunan}
  \country{China}
}

\author{Cheng Tan}
\affiliation{%
  \institution{Westlake University}
  \city{Hangzhou}
  \state{Zhejiang}
  \country{China}
}

\author{Yue Liu}
\author{Ke Liang}
\affiliation{%
  \institution{National University of Defense Technology}
  \city{Changsha}
  \state{Hunan}
  \country{China}
}

\author{Siwei Wang}
\author{Sihang Zhou}
\affiliation{%
  \institution{National University of Defense Technology}
  \city{Changsha}
  \state{Hunan}
  \country{China}
}

\author{Jun Xia}
\author{Stan Z.Li}
\affiliation{%
  \institution{Westlake University}
  \city{Hangzhou}
  \state{Zhejiang}
  \country{China}
}

\author{Xinwang Liu}
\authornote{Corresponding author}
\author{En Zhu}
\authornotemark[1]
\affiliation{%
  \institution{National University of Defense Technology}
  \city{Changsha}
  \state{Hunan}
  \country{China}
}

\renewcommand{\shortauthors}{Xihong Yang et al.}

\begin{abstract}

Contrastive graph node clustering via learnable data augmentation is a hot research spot in the field of unsupervised graph learning. The existing methods learn the sampling distribution of a pre-defined augmentation to generate data-driven augmentations automatically. Although promising clustering performance has been achieved, we observe that these strategies still rely on pre-defined augmentations, the semantics of the augmented graph can easily drift. The reliability of the augmented view semantics for contrastive learning can not be guaranteed, thus limiting the model performance. To address these problems, we propose a novel \textbf{CON}trasti\textbf{V}e Graph Clust\textbf{E}ring network with \textbf{R}eliable Augmen\textbf{T}ation (\textbf{CONVERT}). Specifically, in our method, the data augmentations are processed by the proposed reversible perturb-recover network. It distills reliable semantic information by recovering the perturbed latent embeddings. Moreover, to further guarantee the reliability of semantics, a novel semantic loss is presented to constrain the network via quantifying the perturbation and recovery. Lastly, a label-matching mechanism is designed to guide the model by clustering information through aligning the semantic labels and the selected high-confidence clustering pseudo labels. Extensive experimental results on seven datasets demonstrate the effectiveness of the proposed method. We release the code and appendix of CONVERT at https://github.com/xihongyang1999/CONVERT on GitHub.

\end{abstract}



\maketitle

\begin{figure}[t]
\centering
\includegraphics[scale=0.3]{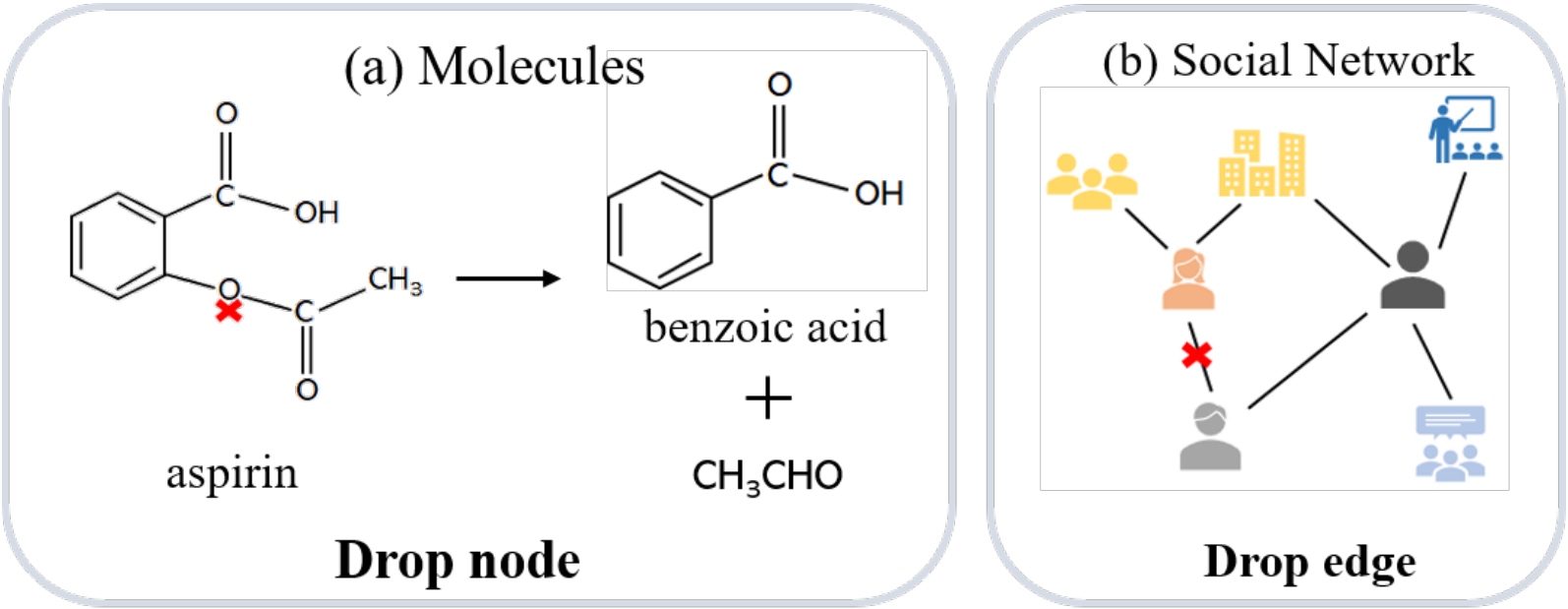}
\caption{Illustration of the semantic drift.}
\label{motivation}  
\end{figure}

\section{INTRODUCTION}

Thanks to the strong representation capacity, graph learning algorithms have attracted more attention in many fields of multimedia, including the recommendation system \cite{chen2_rec,chen3_rec, chen4_rec,huangxu,cor,yq1}, knowledge graph \cite{lk_1,lk_2,lk_3}, temporal graph \cite{lm_1,lm_2,lm_3}, molecular graph \cite{zaixi_1,zaixi_2,zaixi_3,zaixi_4} and so on. Among those directions, graph contrastive clustering \cite{MVGRL,DCRN,GDCL,CCGC,HSAN,AGC-DRR,yuesurvey,Dink_net} has become a hot research spot, which encodes the nodes with graph neural networks into the embeddings and divides them into disjoint clusters in the unsupervised scenario.

In general, prevailing graph contrastive clustering algorithms initially create augmented graph views through perturbations in node attributes or edges. Subsequently, these methods aim to bring identical samples in different views closer together while simultaneously distancing distinct samples from each other. Graph data augmentations are adopted as a crucial technique to construct contrastive views. More recently, learnable graph data augmentation has gained significant attention.

Specifically, through a Bayesian manner, JOAO \cite{JOAO} proposed an augmentation strategy to automatically select augmentations among many pre-defined candidates for graph classification. Although verified effective, the augmentations selected still depend on the pre-defined schemes and are not learnable. To further alleviate this problem, AD-GCL \cite{ADGCL} designed a learnable edge augmentation by Bernoulli distribution. NCLA \cite{NCLA} proposed a learnable topology augmentation method with the multi-head graph attention mechanism. However, the learnable strategy for node level is neglected. Moreover, by acquiring knowledge of a probability distribution, AutoGCL \cite{AutoGCL} introduced an auto-augmentation approach that involves masking or dropping nodes. The model performance is guaranteed by the proposed learnable augmentation strategy. However, previous methods still rely on pre-defined augmentations. The semantics of the augmented view easily drift. As shown in Fig. \ref{motivation} (a), we observe that when dropping nodes in the molecule graph, the semantic information will dramatically change. Similarly, Fig. \ref{motivation} (b) demonstrates that the relationship will be represented incorrectly by dropping the connection in the social network. The semantic reliability of the constructed views can not be guaranteed, thus limiting model clustering performance.

To address these issues, we propose a novel \textbf{CON}trasti\textbf{V}e Graph Clust\textbf{E}ring network with \textbf{R}eliable Augmen\textbf{T}ation \textbf{(CONVERT)}. In our method, a reversible perturb-recover network is designed to generate augmented views by neural network optimization. Concretely, the features of the augmented view are extracted by the perturb and recover operations in the latent space, thus improving the reliability of the semantics. In addition, we propose a novel semantic loss to further guarantee the reliability of semantics by quantifying the perturbation and the recovery of the reversible network. Moreover, the neural networks are guided with clustering information by aligning the selected high-confidence clustering pseudo labels and the semantic labels.

In those manners, we guarantee the reliability of the embedding semantics in dual aspects. Firstly, for the network aspect, we design a reversible perturb-recover network. The recover network restores the embeddings generated by the perturbed network. Thus the semantics of the original view and the augmented view are more similar. Secondly, for the optimization aspect, we design a semantic loss to further guarantee reliability by pushing close the similarity matrix of the embeddings. We summarize the key contributions of this paper as follows:

\begin{itemize}
    \item We propose a contrastive graph clustering method with reliable augmentation, termed CONVERT, by designing a reversible perturb-recover network to generate the augmented view with reliable semantics.
    
    \item To further guarantee the reliability of the semantic, we design a semantic loss by quantifying the perturbation and recovery.
    
    \item In order to guide the model with clustering information, a label-matching mechanism is proposed to align the selected high-confidence pseudo labels and semantic labels.

    \item Extensive experimental results on seven datasets have demonstrated that CONVERT outperforms the existing state-of-the-art deep graph clustering algorithms. Moreover, the effectiveness of the proposed modules is verified by ablation studies.
\end{itemize}

\begin{table}[h]
\centering
\large
\caption{Notation summary.}
\scalebox{0.9}{
\begin{tabular}{@{}ccccccc@{}}
\toprule
 & \textbf{Notations} &  &  & \textbf{Meaning}                         &  &  \\ \midrule
 & $\textbf{X} \in \mathds{R}^{N\times D}$                  &  &  & The Attribute Matrix                     &  &  \\
& $\widetilde{\textbf{X}}_1 \in \mathds{R}^{N\times D}$                  &  &  & The Perturbed Attribute Matrix                      &  &  \\
 & $\textbf{A} \in \mathds{R}^{N\times N}$                  &  &  & The Adjacency Matrix                     &  &  \\
 & $\textbf{D} \in \mathds{R}^{N\times N}$                   &  &  & The Degree Matrix                        &  &  \\
  & $\textbf{E} \in \mathds{R}^{N\times D}$                  &  &  & The Node Embeddings                      &  &  \\
 & $\textbf{H}^{v_2}_p \in \mathds{R}^{N\times D}$                  &  &  & The Perturbed Embeddings                      &  &  \\
 & $ \textbf{H}^{v_1}_r \in \mathds{R}^{N\times N}$                  &  &  & The Recovered Embeddings &  &  \\
 & $\textbf{S} \in \mathds{R}^{N\times N}$                   &  &  & The Similarity Matrix              &  &  \\
  & $\textbf{h} $                   &  &  & High-confidence Clustering Pseudo Label              &  &  \\
 & $\textbf{p}^{se} $                  &  &  & The Semantic Label                    &  &  \\ \bottomrule
\end{tabular}}
\label{NOTATION_TABLE} 
\end{table}

\begin{figure*}
\centering
\scalebox{0.54}{
\includegraphics{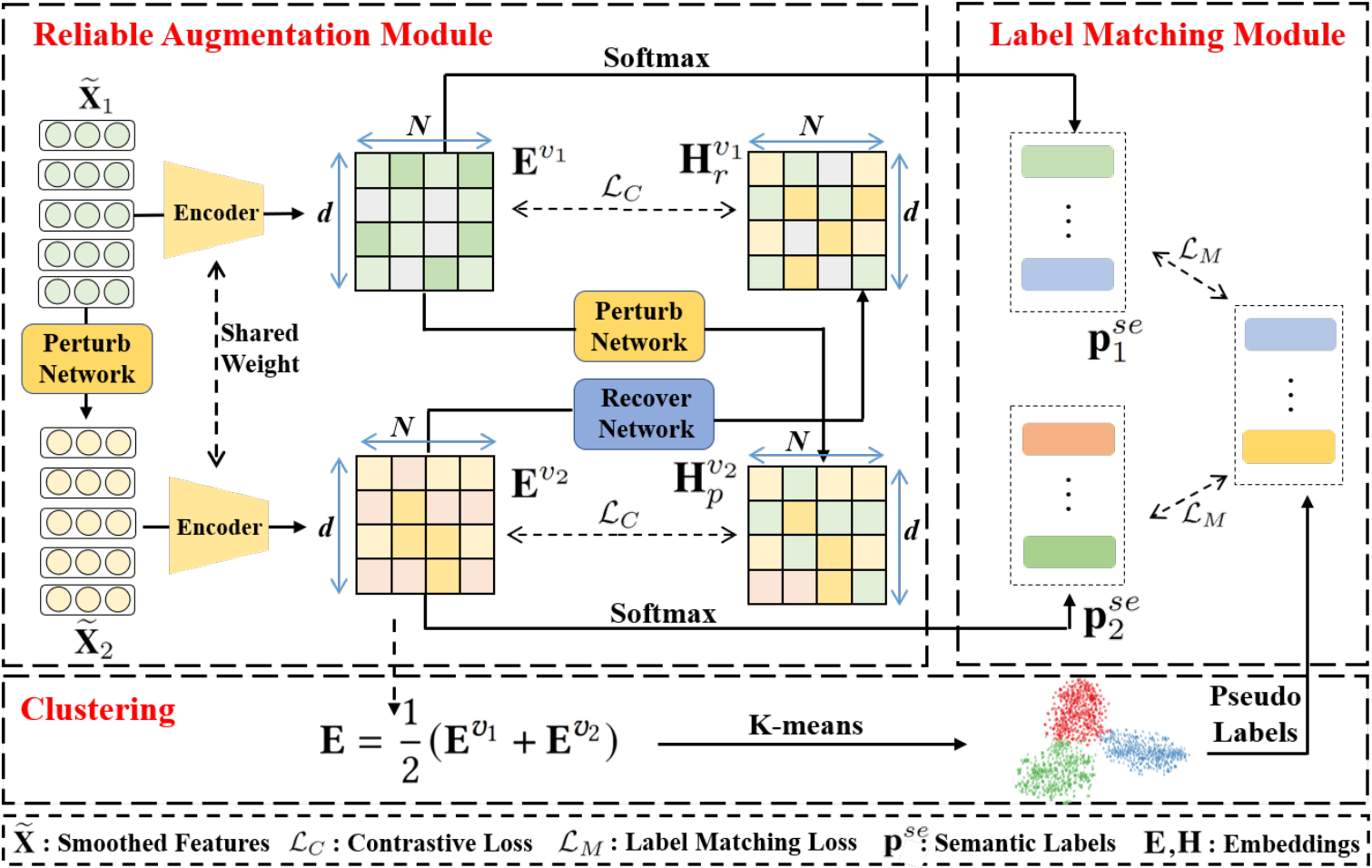}}
\caption{Illustration of CONVERT. In our method, we design a reversible perturb-recover network to generate the augmented view. The semantic information is guaranteed by the perturbation and recovery operation in the latent space. Besides, we design a semantic loss to further improve the reliability of the augmented views. Detailed descriptions are shown in Fig.\ref{keep_semantic}. Lastly, we design a label-matching mechanism to utilize the clustering information. The selected high-confidence clustering pseudo labels align the semantic labels $\textbf{p}^{se}$ with the label matching loss $\mathcal{L}_M$, thus guiding the model to have better performance.}
\label{overall}
\end{figure*}

\section{RELATED WORK}

\subsection{Deep Graph Clustering}
Graph learning methods \cite{yq3, MGCN} have attracted great attention recently. Deep graph node clustering is an important unsupervised downstream task. It aims to learn the graph underlying semantic information and divide the nodes into different clusters. The existing clustering algorithms can be classified into three categories, i.e., generative algorithms \cite{DFCN,DAEGC,SDCN,gc_see}, adversarial algorithms \cite{AGAE,ARGA}, and contrastive algorithms \cite{AGE,MVGRL,DCRN,GDCL,SCGC}. Contrastive learning algorithms have obtained great success in the field of graph \cite{simgrace,GCA,xihong}. In this paper, we mainly focus on contrastive clustering algorithms. Data augmentation plays a crucial role in contrastive clustering algorithms. To be specific, MVGRL \cite{MVGRL}, GDCL \cite{GDCL}, and DCRN \cite{DCRN} utilize the graph diffusion matrix as the augmented view. Different from the above algorithms, SCAGC \cite{SCAGC} conducts random edge perturbation to construct the augmented view. Regarding feature operation, both DCRN and SCAGC execute augmentations on node attributes through attribute corruption. Despite their proven effectiveness, the strong clustering performance of these techniques is intricately tied to the judicious selection of data augmentations. In recent developments, certain graph augmentation methodologies \cite{AFGRL} emphasize that specific data augmentations may trigger semantic drift. To alleviate this problem, in our paper, we design a reversible network to generate the augmented view in a perturb-recover way. The semantics of embeddings are guaranteed by the reversible network, thus avoiding semantic drift.

\subsection{Graph Data Augmentation}

Graph data augmentation \cite{sail,wangaug1,wangaug2} has emerged as a dominant technique in graph contrastive learning. The existing data augmentation methods could roughly be divided into three classes. 1) Augment-free methods. AFGRL \cite{AFGRL} devises the augmented view by identifying nodes equipped with local and global data, eschewing the need for augmentation. However, this approach doesn't assure the augmented view's reliability, potentially resulting in subpar performance. 2) Adaptive augmentation techniques. In the realm of graph classification, JOAO \cite{JOAO} harnesses the potential to autonomously choose data augmentation through learning the sampling distribution of predefined augmentations. Furthermore, GCA \cite{GCA} enhances augmentation adaptability by integrating diverse priors targeting the graph's topological and semantic attributes. However, the augmentation is still not optimized by the neural network in the adaptive augmentation methods. 3) Learnable data augmentation. An edge-level learnable strategy is designed in AD-GCL \cite{ADGCL} while neglecting the augmentations on the node level. Similarly, NCLA \cite{NCLA} proposed a learnable topology by the multi-head graph attention mechanism. However, those two methods neglected the learnable strategy for the node level. Subsequently, AutoGCL \cite{AutoGCL} introduced a probability-driven learnable strategy. Despite yielding improved the model performance, AutoGCL remains reliant on existing and predefined data augmentations. In contrast, our work introduces a learnable augmentation approach at the embedding level. Through a reversible network, we generate the augmented view, allowing the network to optimize the view's quality during training.

\section{Method}

In this section, we introduce a novel Contrastive Graph Clustering method with Reliable Augmentation (CONVERT). The comprehensive CONVERT framework is illustrated in Figure \ref{overall}. Primarily, CONVERT encompasses two key modules: the Learnable Augmentation Module with Reliable Augmentation and the Label-Matching Module. Detailed definitions for these modules will be presented in subsequent sections.

\subsection{Notations}

For an undirected graph $\textbf{G}=\left \{\textbf{X}, \textbf{A} \right \}$, consider ${\textbf{V}}=\{v_1, v_2, \dots, v_N\}$ as a set of $N$ nodes categorized into $K$ classes. Here, $\textbf{X} \in \mathds{R}^{N\times D}$ denotes the attribute matrix, while $\textbf{A} \in \mathds{R}^{N\times N}$ stands for the attribute matrix and the original adjacency matrix. The degree matrix is symbolized as $\textbf{D}=diag(d_1, d_2, \dots ,d_N)\in \mathds{R}^{N\times N}$, where $d_i=\sum_{(v_i,v_j)\in \mathcal{E}}a_{ij}$. The graph Laplacian matrix is defined as $\textbf{L}=\textbf{D}-\textbf{A}$. Utilizing the renormalization technique $\widehat{\textbf{A}} = \textbf{A} + \textbf{I}$ from GCN \cite{GCN}, the symmetric normalized graph Laplacian matrix is represented as $\widetilde{\textbf{L}} = \textbf{I} - \widehat{\textbf{D}}^{-\frac{1}{2}}\widehat{\textbf{A}}\widehat{\textbf{D}}^{-\frac{1}{2}}$. A summary of essential notations is provided in Table \ref{NOTATION_TABLE}.

\subsection{Reliable Augmentation Module}

In this subsection, we design a reversible network to obtain the augmented views with reliable semantics. Specifically, we propose a perturb network and a recover network.

To avoid the entanglement of graph convolutional filters and weight matrices, following AGE \cite{AGE}, we firstly adopt the Laplacian filter to obtain the smoothed attribute matrix $\widetilde{\textbf{X}}_1$ as follows: 

\begin{equation} 
\widetilde{\textbf{X}}_1 = (\textbf{I}-\widetilde{\textbf{L}})^\text{t}\textbf{X},
\label{filter}
\end{equation}
where $\widetilde{\textbf{L}}$ represents the symmetric normalized graph Laplacian matrix, while $\text{t}$ denotes the layer number of the Laplacian filter. Subsequently, we leverage the perturbation network $p(\cdot; \theta)$ to apply perturbations to the attribute matrix, leading to the formation of the perturbed attribute matrix $\widetilde{\textbf{X}}_2$, formulated as follows:

\begin{equation} 
\widetilde{\textbf{X}}_2 = p (\widetilde{\textbf{X}}_1;\theta).
\label{perturbed_smoothed_feature}
\end{equation}

After that, we extract the original embeddings $\textbf{E}^{v_1}$ and the perturbed embeddings $\textbf{E}^{v_2}$ for the $\widetilde{\textbf{X}}_1$ and $\widetilde{\textbf{X}}_2$ with the encoder network:

\begin{equation} 
\begin{aligned}
\textbf{E}^{v_1} &= \text{Encoder}(\widetilde{\textbf{X}}_1); \textbf{E}^{v_1}_i = \frac{\textbf{E}^{v_1}_i}{||\textbf{E}^{v_1}_i||_2}, i = 1,2,\dots, N \\
\textbf{E}^{v_2} &= \text{Encoder}(\widetilde{\textbf{X}}_2); \textbf{E}^{v_2}_j = \frac{\textbf{E}^{v_2}_j}{||\textbf{E}^{v_2}_j||_2}, j = 1,2,\dots, N. 
\label{encoder}
\end{aligned}
\end{equation}

We adopt the multi-layers perceptions (MLPs) as the encoder network. To obtain the reliable semantic augmented views, we design a recover network $r(\cdot~;\theta)$ to recover the perturbed embeddings as follows:
\begin{equation} 
\textbf{H}^{v_1}_r = r (\textbf{E}^{v_2};\theta),
\label{perturbed_smoothed_feature}
\end{equation}
where $\textbf{H}^{v_1}_r$ denotes the recovered embeddings. Here, we adopt the two layers of multi-layers perceptions as the perturb network and the recover network, respectively. In this manner, the perturbed embeddings could be recovered by $r(\cdot~;\theta)$. In embedding space, the recover network restores the perturbed semantics. Therefore, the recovered embeddings have more similar semantics to the original embeddings. In addition, we implement the perturb to the original embeddings $\textbf{E}^{v_1}$ by the perturb network $p(\cdot~;\theta)$ as formulated:

\begin{equation} 
\textbf{H}^{v_2}_p = p (\textbf{E}^{v_1};\theta),
\label{perturbed_origin_embeddings}
\end{equation}
where $\textbf{H}^{v_2}_p$ is the perturbed embeddings. Since the smoothed attributed matrix $\widetilde{\textbf{X}}$ has been perturbed, we conduct the same operation for the original embeddings to obtain the perturbed embeddings. In this way, the semantics of the perturbed embeddings are similar to the same perturbed operation in latent space.

\begin{figure}
\centering
\scalebox{0.35}{
\includegraphics{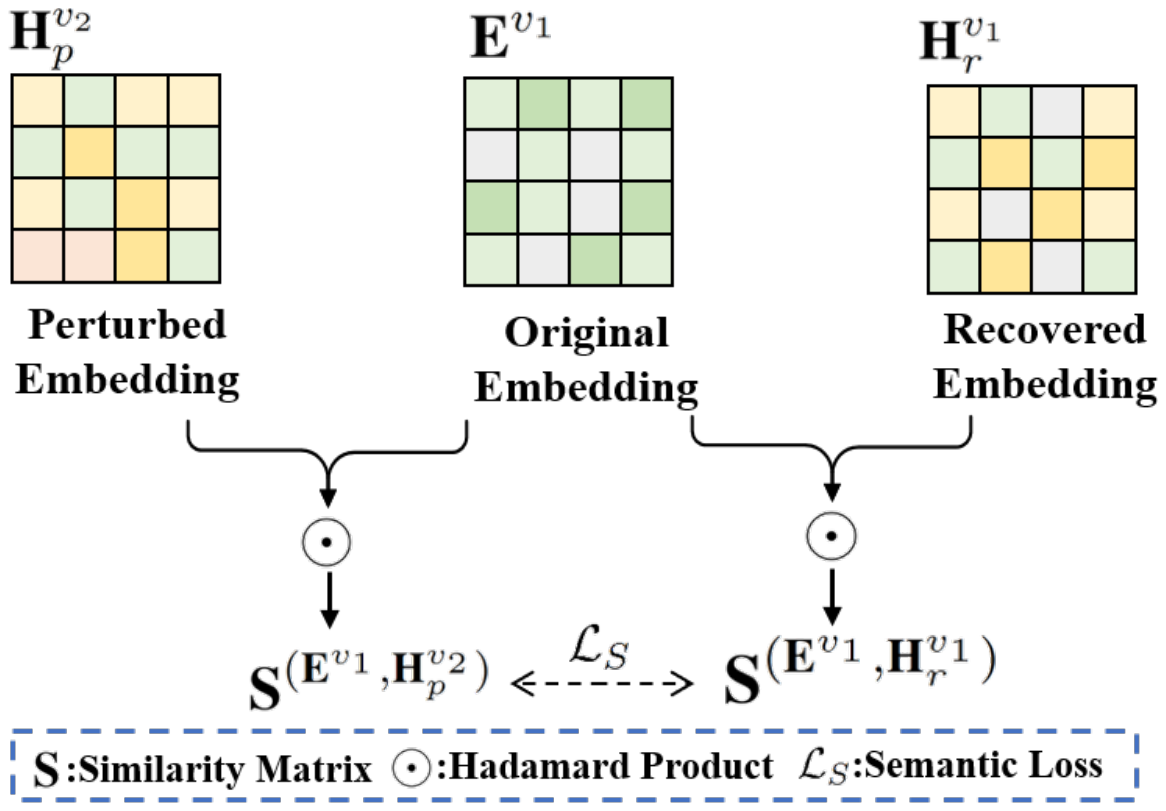}}
\caption{Illustration of the semantic loss $\mathcal{L}_S$. Through pulling close $\textbf{S}^{(\textbf{E}^{v_1}, \textbf{H}^{v_1}_r)}$ and $ \textbf{S}^{( \textbf{E}^{v_1},  \textbf{H}^{v_2}_p)}$, the perturbation and recovery could be quantified, thus guaranteeing the reliability.}
\label{keep_semantic}
\end{figure}

Aiming to further guarantee reliable semantics of the augmented view, as shown in Fig.~\ref{keep_semantic}, we calculate the similarity matrix $\textbf{S}$ between $i$-th sample in the first view and $j$-th sample in the second view as follows:

\begin{equation} 
\begin{aligned}
 \textbf{S}^{(\textbf{E}^{v_1},  \textbf{H}^{v_1}_r)}_{ij} &= \textbf{E}^{v_1}_{i} \odot~{\textbf{H}^{v_1}_{rj}},\\
 \textbf{S}^{( \textbf{E}^{v_1},  \textbf{H}^{v_2}_p)}_{ij} &= \textbf{E}^{v_1}_i \odot~ { \textbf{H}^{v_2}_{pj}},\\
\end{aligned}
\label{sim_semantic}
\end{equation}
where $\odot$ denotes the Hadamard product. Besides, $\textbf{S}^{(\textbf{E}^{v_1},  \textbf{H}^{v_1}_r)}$ represents the similarity matrix between the original embeddings and the recovered embeddings. $ \textbf{S}^{( \textbf{E}^{v_1},  \textbf{H}^{v_2}_p)}$ denotes the similarity matrix of the original embeddings and the perturbed embeddings. $\textbf{S}$ can better reveal the changing of semantics by quantifying perturbation and recovery. Subsequently, we enhance the reliability of embedding semantics by forcing the similarity matrix to pull close:  

\begin{equation} 
\mathcal{L}_S = \left \| \textbf{S}^{(\textbf{E}^{v_1},  \textbf{H}^{v_1}_r)} - \textbf{S}^{( \textbf{E}^{v_1},  \textbf{H}^{v_2}_p)} \right \|^2_F.
\label{semantic_loss}
\end{equation}

Through Eq.\ref{semantic_loss}, we constrain the network to restore the perturbed embeddings. In this manner, the quantification of perturb and restore is mutually reinforcing, thus generating the augmented view with more reliable semantics. 

Different from previous augmentation methods, our method can construct more reliable semantic augmented views. The reasons are as follows. On the one hand, instead of using normal graph augmentations, e.g., feature masking or edge perturbing, the augmented views are generated by the reversible network. The process of the augmented view is learnable and can be optimized by the network. On the other hand, the semantics of the augmented view is guaranteed by the perturb-recover procedure. The perturbed embeddings are restored through the recover network.

\begin{algorithm}[t]
\small
\caption{\textbf{CONVERT}}
\label{ALGORITHM}
\flushleft{\textbf{Input}: The input graph $\textbf{G}=\{\textbf{X},\textbf{A}\}$; The iteration number $I$; High-confidence epoch $N$} \\
\flushleft{\textbf{Output}: The clustering result \textbf{R}.} 
\begin{algorithmic}[1]
\FOR{$i=1$ to $I$}
\STATE Acquire the smoothed attribute matrix $\widetilde{\textbf{X}}_1$ and perturbed attribute matrix $\widetilde{\textbf{X}}2$.
\STATE Encoder the attribute to obtain the embeddings $\textbf{E}^{v_1}$ and $\textbf{E}^{v_2}$.
\STATE Obtain the perturbed embedding $\textbf{H}^{v_2}_p$ and the recovered embedding $\textbf{H}^{v_1}_r$ with Eq.\eqref{perturbed_smoothed_feature} and \eqref{perturbed_origin_embeddings}.
\STATE Fuse $\textbf{E}^{v_1}$ and $\textbf{E}^{v_2}$ to to yield $\textbf{E}$, followed by K-means for clustering results.
\STATE Calculate the similarity matrix using Eq.\eqref{sim_semantic}.
\STATE Obtain the semantic labels $\textbf{p}^{se}$ and high-confidence pseudo labels $\textbf{h}$ via Eq.\eqref{semantic_label} and \eqref{high_labels}.
\IF{$i>N$}
\STATE Calculate the label-matching loss $\mathcal{L}_{M}$ via Eq.\eqref{match_loss}
\ENDIF
\STATE Calculate the semantic loss $\mathcal{L}_{S}$ with Eq. \eqref{semantic_loss}.
\STATE Calculate the contrastive loss $\mathcal{L}_{c}$ with Eq. \eqref{simclr_loss}.
\STATE Update the entire network by minimizing $\mathcal{L}$ in Eq. \eqref{total_Loss}.
\ENDFOR
\STATE Perform K-means on \textbf{E} to obtain the final clustering result \textbf{R}. 
\STATE \textbf{return} \textbf{R}
\end{algorithmic}
\end{algorithm}

\subsection{Label-Matching Module}

After encoding, we firstly fuse the embeddings to obtain consensus embeddings $\textbf{E}$ as below:

\begin{equation}
\textbf{E} = \frac{1}{2}(\textbf{E}^{v_1}+\textbf{E}^{v_2}).
\label{fuse}
\end{equation}

Then we implement K-means \cite{KMEANS} to obtain clustering results. To utilize more reliable clustering information, we obtain the high-confidence pseudo labels $\textbf{h}$ through selecting clustering pseudo labels $\textbf{p}$. Formally,

\begin{equation}
\textbf{h} = \text{top} (\textbf{p}),
\label{high_labels}
\end{equation}
where $\text{top}(\cdot)$ is the confidence measure function to select the top $\tau$ clustering pseudo labels.

In the unsupervised clustering scene, the supervision information could be extracted by the high-confidence clustering pseudo labels. We design a matching mechanism between the semantic labels and the high-confidence pseudo labels to further guide the network. Specifically, we conduct the Softmax for the embeddings to obtain the semantic labels $\textbf{p}^{se}$ as follows:

\begin{equation}
\textbf{p}^{se}_i = \text{Softmax}(\textbf{E}^{v_i}), i \in {1,2}.
\label{semantic_label}
\end{equation}

Subsequently, we match the semantic labels $\textbf{p}^{se}$ and the high-confidence pseudo labels $\textbf{h}$ by:
\begin{equation}
\mathcal{L}_M = \text{CE} (\textbf{p}^{se}_i, \textbf{h}), i \in {1,2}, 
\label{match_loss}
\end{equation}
where $\text{CE}(\cdot)$ is the Cross-Entropy loss \cite{cross-entropy}. Through the matching mechanism, the quality of the semantics can be further improved with the high-confidence pseudo labels. Furthermore, in order to enhance the precision of clustering pseudo labels, we employ a two-stage training strategy aimed at bolstering the network's discriminative capability. Specifically, during the second stage, we opt for high-confidence pseudo labels to enact the matching mechanism.

\subsection{Loss Function}

The proposed CONVERT jointly optimizes three objectives, including the semantic loss $\mathcal{L}_S$, the label matching loss $\mathcal{L}_M$, and the contrastive loss $\mathcal{L}_{C}$. More precisely, we employ $\mathcal{L}_{C}$ to amplify the similarity among positive samples while diminishing it among negative samples. The contrastive loss can be articulated as follows:

\begin{equation}
\begin{aligned}
\ell(i) &= -\log\frac{\text{e}^{\text{sim}(\textbf{E}_i^{v_j}, \textbf{E}_i^{v_l})}}{\sum_{k = 1,k\neq i}^N{\text{e}^{\text{sim} (\textbf{E}_i^{v_j}, \textbf{E}_k^{v_l})}}}, j,l \in \{1,2\}\\
\mathcal{L}_C &= \frac{1}{N}{\sum_{i=1}^{N} \ell(i)}
\end{aligned}
\label{simclr_loss}
\end{equation}
where $\text{sim}(\cdot)$ represents the function to calculate the similarity, i.e., cosine similarity. Here, we calculate the contrastive loss for the view pairs $(\textbf{E}^{v_1},\textbf{H}^{v_1}_r)$ and $(\textbf{E}^{v_2},\textbf{H}^{v_2}_p)$. In summary, the objective of CONVERT is formulated as:
\begin{equation}
\mathcal{L} = \mathcal{L}_{C}+ \alpha \mathcal{L}_{S} + \beta \mathcal{L}_M,
\label{total_Loss}
\end{equation}
where $\alpha$ and $\beta$ are the trade-off hyper-parameters. The detailed learning procedure of CONVERT is illustrated in Algorithm \ref{ALGORITHM}.

\section{EXPERIMENT}

In this section, we conduct the experiments to verify the effectiveness of the proposed CONVERT through answering the following questions:

\begin{itemize}
\item \textbf{RQ1}: How effective is CONVERT for graph node clustering?
\item \textbf{RQ2}: How does the proposed module influence the performance of CONVERT?
\item \textbf{RQ3}: How about the efficient about CONVERT ?
\item \textbf{RQ4}: How do the hyper-parameters impact the performance of CONVERT?
\item \textbf{RQ5}: What is the clustering structure revealed by CONVERT?
\end{itemize}

\begin{table}[t]
\centering
\caption{Statistics summary of seven datasets.}
\scalebox{0.9}{
\begin{tabular}{@{}cccccc@{}}
\toprule
\textbf{Dataset} & \textbf{Type} & \textbf{Sample} & \textbf{Dimension} & \textbf{Edge}  & \textbf{Class} \\ \midrule
\textbf{CORA}  & Graph   & 2708    & 1433       & 5429   & 7       \\
\textbf{CITESEER} & Graph    & 3327    & 3703      & 4732   & 6       \\
\textbf{AMAP} & Graph  & 7650   & 745       & 119081  & 8       \\
\textbf{CORAFULL} & Graph  & 19793   & 8710       & 63421  & 70       \\
\textbf{BAT}    & Graph  & 131    & 81      & 1038  & 4       \\
\textbf{EAT}    & Graph & 399    & 203       & 5994 & 4       \\
\textbf{UAT} & Graph  & 1190   & 239       & 13599  & 4       \\\bottomrule
\end{tabular}}
\label{DATASET_INFO} 
\end{table}

\begin{table*}[]
\centering
\caption{The clustering performance is gauged through ten runs, encompassing mean values and standard deviations. Notably, the most exceptional and second-best outcomes are denoted by {\color[HTML]{FF0000}red} and {\color[HTML]{0000FF}blue} values correspondingly. "OOM" signifies out-of-memory during training.}
\setlength{\tabcolsep}{4pt}
\scalebox{0.7}{
\begin{tabular}{c|c|ccc|ccc|ccccc|c}
\hline
                                    &                                   & \multicolumn{3}{c|}{\textbf{Classical Graph Clustering Methods}}        & \multicolumn{3}{c|}{\textbf{Constrastive Clustering Methods}}                 & \multicolumn{5}{c|}{\textbf{Graph Augmentation Methods}}                                                                                                              & \textbf{}                         \\ \cline{3-14} 
                                    &                                   & \textbf{DAEGC}    & \textbf{ARGA}                     & \textbf{SDCN}   & \textbf{AGE}                      & \textbf{MVGRL}   & \textbf{AGC-DRR}                  & \textbf{GCA}                      & \textbf{AFGRL}        & \textbf{AutoSSL}                  & \textbf{SUBLIME}                  & \textbf{NACL}                     & \textbf{CONVERT}                    \\
\multirow{-3}{*}{\textbf{Dataset}}  & \multirow{-3}{*}{\textbf{Metric}} & \textbf{IJCAI 19} & \textbf{TCYB 19}                  & \textbf{WWW 20} & \textbf{SIGKDD 20}                & \textbf{ICML 20} & \textbf{IJCAI 22}                 & \textbf{WWW 21}                   & \textbf{AAAI 22}      & \textbf{ICLR 22}                  & \textbf{WWW 22}                   & \textbf{AAAI 23}                  & \textbf{Ours}                     \\ \hline
                                    & \textbf{ACC}                      & 70.43±0.36        & 71.04±0.25                        & 35.60±2.83      & {\color[HTML]{0000FF} 73.50±1.83} & 70.47±3.70       & 40.62±0.55                        & 53.62±0.73                        & 26.25±1.24            & 63.81±0.57                        & 71.14±0.74                        & 51.09±1.25                        & {\color[HTML]{FF0000} 74.07±1.51} \\
                                    & \textbf{NMI}                      & 52.89±0.69        & 51.06±0.52                        & 14.28±1.91      & {\color[HTML]{FF0000} 57.58±1.42} & 55.57±1.54       & 18.74±0.73                        & 46.87±0.65                        & 12.36±1.54            & 47.62±0.45                        & 53.88±1.02                        & 31.80±0.78                        & {\color[HTML]{0000FF} 55.57±1.12} \\
                                    & \textbf{ARI}                      & 49.63±0.43        & 47.71±0.33                        & 07.78±3.24      & 50.10±2.14                        & 48.70±3.94       & 14.80±1.64                        & 30.32±0.98                        & 14.32±1.87            & 38.92±0.77                        & {\color[HTML]{0000FF} 50.15±0.14} & 36.66±1.65                        & {\color[HTML]{FF0000} 50.58±2.01} \\
\multirow{-4}{*}{\textbf{CORA}}     & \textbf{F1}                       & 68.27±0.57        & 69.27±0.39                        & 24.37±1.04      & {\color[HTML]{0000FF} 69.28±1.59} & 67.15±1.86       & 31.23±0.57                        & 45.73±0.47                        & 30.20±1.15            & 56.42±0.21                        & 63.11±0.58                        & 51.12±1.12                        & {\color[HTML]{FF0000} 72.92±3.27} \\ \hline
                                    & \textbf{ACC}                      & 75.96±0.23        & 69.28±2.30                        & 53.44±0.81      & 75.98±0.68                        & 41.07±3.12       & {\color[HTML]{0000FF} 76.81±1.45} & 56.81±1.44                        & 75.51±0.77            & 54.55±0.97                        & 27.22±1.56                        & 67.18±0.75                        & {\color[HTML]{FF0000} 77.19±0.55} \\
                                    & \textbf{NMI}                      & 65.25±0.45        & 58.36±2.76                        & 44.85±0.83      & 65.38±0.61                        & 30.28±3.94       & {\color[HTML]{0000FF} 66.54±1.24} & 48.38±2.38                        & 64.05±0.15            & 48.56±0.71                        & 06.37±1.89                        & 63.63±1.07                        & {\color[HTML]{FF0000} 67.20±1.07} \\
                                    & \textbf{ARI}                      & 58.12±0.24        & 44.18±4.41                        & 31.21±1.23      & 55.89±1.34                        & 18.77±2.34       & {\color[HTML]{0000FF} 60.15±1.56} & 26.85±0.44                        & 54.45±0.48            & 26.87±0.34                        & 05.36±2.14                        & 46.30±1.59                        & {\color[HTML]{FF0000} 60.79±1.83} \\
\multirow{-4}{*}{\textbf{AMAP}}     & \textbf{F1}                       & 69.87±0.54        & 64.30±1.95                        & 50.66±1.49      & 71.74±0.93                        & 32.88±5.50       & 71.03±0.64                        & 53.59±0.57                        & 69.99±0.34            & 54.47±0.83                        & 15.97±1.53                        & {\color[HTML]{0000FF} 73.04±1.08} & {\color[HTML]{FF0000} 74.03±1.00} \\ \hline
                                    & \textbf{ACC}                      & 52.67±0.00        & {\color[HTML]{0000FF} 67.86±0.80} & 53.05±4.63      & 56.68±0.76                        & 37.56±0.32       & 47.79±0.02                        & 54.89±0.34                        & 50.92±0.44            & 42.43±0.47                        & 45.04±0.19                        & 47.48±0.64                        & {\color[HTML]{FF0000} 78.02±1.36} \\
                                    & \textbf{NMI}                      & 21.43±0.35        & {\color[HTML]{0000FF} 49.09±0.54} & 25.74±5.71      & 36.04±1.54                        & 29.33±0.70       & 19.91±0.24                        & 38.88±0.23                        & 27.55±0.62            & 17.84±0.98                        & 22.03±0.48                        & 24.36±1.54                        & {\color[HTML]{FF0000} 53.54±1.71} \\
                                    & \textbf{ARI}                      & 18.18±0.29        & {\color[HTML]{0000FF} 42.02±1.21} & 21.04±4.97      & 26.59±1.83                        & 13.45±0.03       & 14.59±0.13                        & 26.69±2.85                        & 21.89±0.74            & 13.11±0.81                        & 14.45±0.87                        & 24.14±0.98                        & {\color[HTML]{FF0000} 51.95±2.18} \\
\multirow{-4}{*}{\textbf{BAT}}      & \textbf{F1}                       & 52.23±0.03        & {\color[HTML]{0000FF} 67.02±1.15} & 46.45±5.90      & 55.07±0.80                        & 29.64±0.49       & 42.33±0.51                        & 53.71±0.34                        & 46.53±0.57            & 34.84±0.15                        & 44.00±0.62                        & 42.25±0.34                        & {\color[HTML]{FF0000} 77.77±1.48} \\ \hline
                                    & \textbf{ACC}                      & 36.89±0.15        & {\color[HTML]{0000FF} 52.13±0.00} & 39.07±1.51      & 47.26±0.32                        & 32.88±0.71       & 37.37±0.11                        & 48.51±1.55                        & 37.42±1.24            & 31.33±0.52                        & 38.80±0.35                        & 36.06±1.24                        & {\color[HTML]{FF0000} 58.35±0.18} \\
                                    & \textbf{NMI}                      & 05.57±0.06        & 22.48±1.21                        & 08.83±2.54      & 23.74±0.90                        & 11.72±1.08       & 07.00±0.85                        & {\color[HTML]{0000FF} 28.36±1.23} & 11.44±1.41            & 07.63±0.85                        & 14.96±0.75                        & 21.46±1.80                        & {\color[HTML]{FF0000} 33.36±0.16} \\
                                    & \textbf{ARI}                      & 05.03±0.08        & 17.29±0.50                        & 06.31±1.95      & 16.57±0.46                        & 04.68±1.30       & 04.88±0.91                        & 19.61±1.25                        & 06.57±1.73            & 02.13±0.67                        & 10.29±0.88                        & {\color[HTML]{0000FF} 21.48±0.64} & {\color[HTML]{FF0000} 27.11±0.19} \\
\multirow{-4}{*}{\textbf{EAT}}      & \textbf{F1}                       & 34.72±0.16        & {\color[HTML]{0000FF} 52.75±0.07} & 33.42±3.10      & 45.54±0.40                        & 25.35±0.75       & 35.20±0.17                        & 48.22±0.33                        & 30.53±1.47            & 21.82±0.98                        & 32.31±0.97                        & 31.25±0.96                        & {\color[HTML]{FF0000} 58.42±0.22} \\ \hline
                                    & \textbf{ACC}                      & 52.29±0.49        & 49.31±0.15                        & 52.25±1.91      & {\color[HTML]{0000FF} 52.37±0.42} & 44.16±1.38       & 42.64±0.31                        & 39.39±1.46                        & 41.50±0..25           & 42.52±0.64                        & 48.74±0.54                        & 45.38±1.15                        & {\color[HTML]{FF0000} 57.36±0.55} \\
                                    & \textbf{NMI}                      & 21.33±0.44        & {\color[HTML]{0000FF} 25.44±0.31} & 21.61±1.26      & 23.64±0.66                        & 21.53±0.94       & 11.15±0.24                        & 24. 05±0.25                       & 17.33±0.54            & 17.86±0.22                        & 21.85±0.62                        & 24.49±0.57                        & {\color[HTML]{FF0000} 28.75±1.13} \\
                                    & \textbf{ARI}                      & 20.50±0.51        & 16.57±0.31                        & 21.63±1.49      & 20.39±0.70                        & 17.12±1.46       & 09.50±0.25                        & 14. 37±0.19                       & 13.62±0.57            & 13.13±0.71                        & 19.51±0.45                        & {\color[HTML]{0000FF} 21.34±0.78} & {\color[HTML]{FF0000} 27.96±0.79} \\
\multirow{-4}{*}{\textbf{UAT}}      & \textbf{F1}                       & 50.33±0.64        & 50.26±0.16                        & 45.59±3.54      & 50.15±0.73                        & 39.44±2.19       & 35.18±0.32                        & 35.72±0.28                        & 36.52±0.89            & {\color[HTML]{0000FF} 52.94±0.87} & 46.19±0.87                        & 30.56±0.25                        & {\color[HTML]{FF0000} 54.55±1.49} \\ \hline
                                    & \textbf{ACC}                      & 64.54±1.39        & 61.07±0.49                        & 65.96±0.31      & {\color[HTML]{FF0000} 68.73±0.24}                        & 62.83±1.59       & {\color[HTML]{0000FF} 68.32±1.83} & 60.45±1.03                        & 31.45±0.54            & 66.76±0.67                        & 64.14±0.65                        & 59.23±2.32                        & {68.43±0.69} \\
                                    & \textbf{NMI}                      & 36.41±0.86        & 34.40±0.71                        & 38.71±0.32      & {\color[HTML]{FF0000} 44.93±0.53} & 40.69±0.93       & 40.28±1.41                        & 36.15±0.78                        & 15.17±0.47            & 40.67±0.84                        & 39.08±0.25                        & 36.68±0.89                        & {\color[HTML]{0000FF} 41.62±0.73} \\
                                    & \textbf{ARI}                      & 37.78±1.24        & 34.32±0.70                        & 40.17±0.43      & {\color[HTML]{0000FF} 45.31±0.41} & 34.18±1.73       & {\color[HTML]{FF0000} 45.34±2.33} & 35.20±0.96                        & 14.32±0.78            & 38.73±0.55                        & 39.27±0.78                        & 33.37±0.53                        & 42.77±1.63                        \\
\multirow{-4}{*}{\textbf{CITESEER}} & \textbf{F1}                       & 62.20±1.32        & 58.23±0.31                        & 63.62±0.24      & {\color[HTML]{0000FF} 64.45±0.27}                        & 59.54±2.17       & {\color[HTML]{FF0000} 64.82±1.60} & 56.42±0.94                        & 30.20±0.71            & 58.22±0.68                        & 61.00±0.15                        & 52.67±0.64                        & {62.39±2.15} \\ \hline
                                    & \textbf{ACC}                      & 34.35±1.00        & 22.07±0.43                        & 26.67±0.40      & {\color[HTML]{0000FF} 39.62±0.13} & 31.52±2.95       &                                   & 31.19±0.57                        &                       &                                   & 32.63±1.24                        &                                   & {\color[HTML]{FF0000} 43.53±0.96} \\
                                    & \textbf{NMI}                      & 49.16±0.73        & 41.28±0.25                        & 37.38±0.39      & {\color[HTML]{0000FF} 52.38±0.17} & 48.99±3.95       &                                   & 50.23±1.54                        &                       &                                   & 38.46±0.25                        &                                   & {\color[HTML]{FF0000} 54.86±0.33} \\
                                    & \textbf{ARI}                      & 22.60±0.47        & 12.38±0.24                        & 13.63±0.27      & 24.46±0.48                        & 19.11±2.63       &                                   & 19.17±0.47                        &                       &                                   & {\color[HTML]{FF0000} 31.41±0.55} &                                   & {\color[HTML]{0000FF} 30.39±1.04} \\
\multirow{-4}{*}{\textbf{CORAFULL}} & \textbf{F1}                       & 26.96±1.33        & 18.85±0.41                        & 22.14±0.43      & {\color[HTML]{0000FF} 31.22±0.87} & 26.51±2.87       & \multirow{-4}{*}{OOM}             & 25.44±0.99                        & \multirow{-4}{*}{OOM} & \multirow{-4}{*}{OOM}             & 28.94±1.72                        & \multirow{-4}{*}{OOM}             & {\color[HTML]{FF0000} 35.58±0.98} \\ \hline
\end{tabular}}
\label{compare_result}
\end{table*}

\begin{figure*}[!t]
\footnotesize
\begin{minipage}{0.139\linewidth}
\centerline{\includegraphics[width=\textwidth]{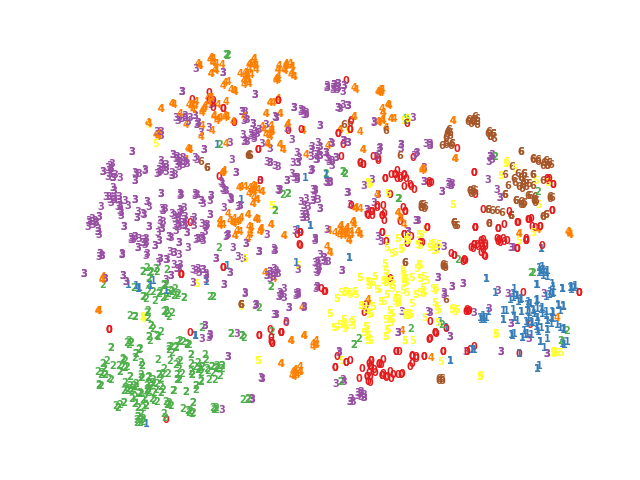}}
\vspace{3pt}
\centerline{\includegraphics[width=\textwidth]{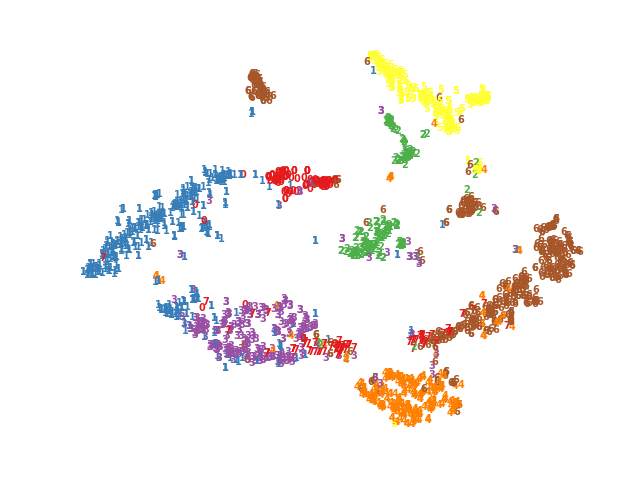}}
\vspace{3pt}
\centerline{DAEGC}
\end{minipage}
\begin{minipage}{0.139\linewidth}
\centerline{\includegraphics[width=\textwidth]{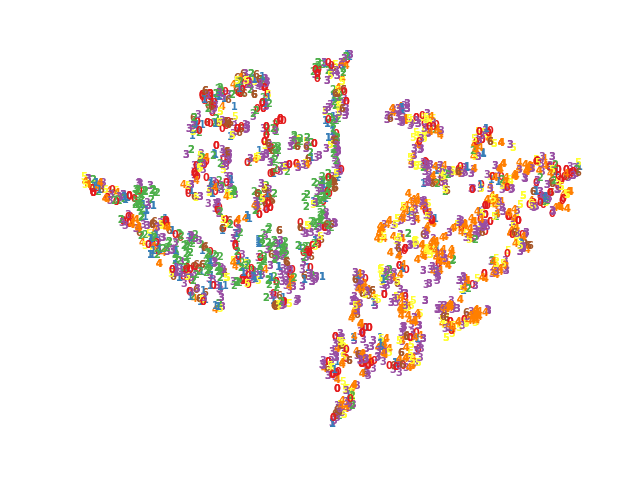}}
\vspace{3pt}
\centerline{\includegraphics[width=\textwidth]{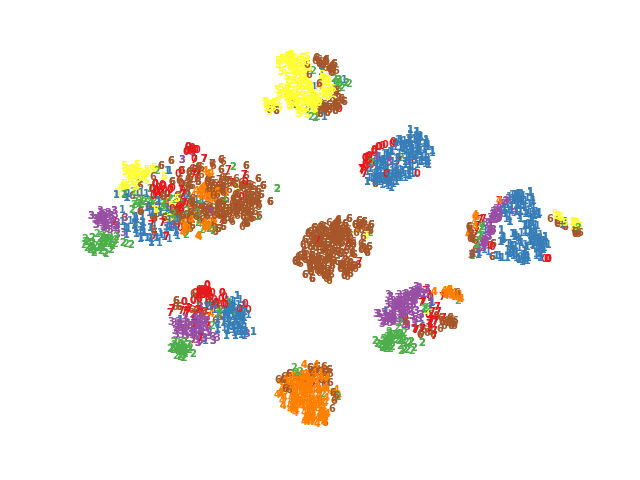}}
\vspace{3pt}
\centerline{SDCN}
\end{minipage}
\begin{minipage}{0.139\linewidth}
\centerline{\includegraphics[width=\textwidth]{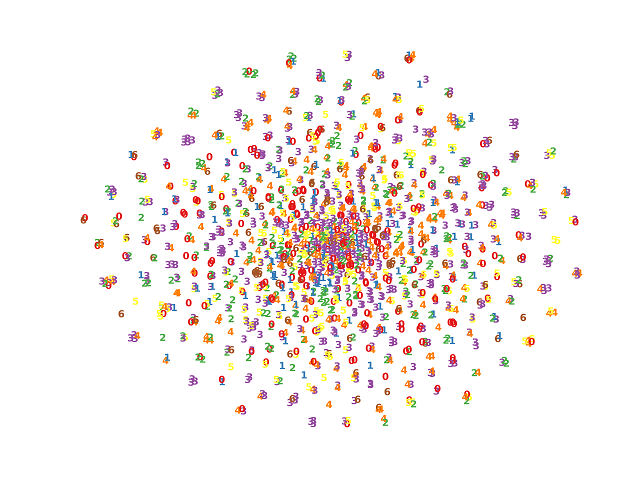}}
\vspace{3pt}
\centerline{\includegraphics[width=\textwidth]{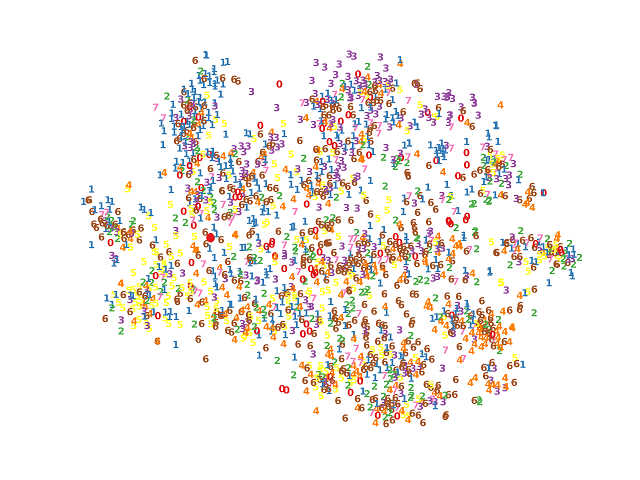}}
\vspace{3pt}
\centerline{AFGRL}
\end{minipage}
\begin{minipage}{0.139\linewidth}
\centerline{\includegraphics[width=\textwidth]{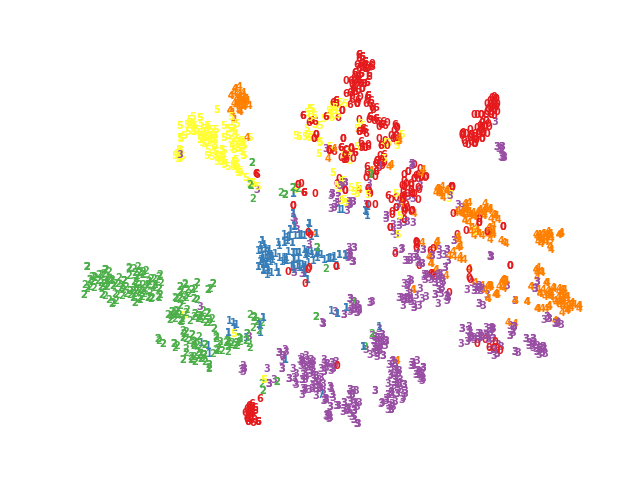}}
\vspace{3pt}
\centerline{\includegraphics[width=\textwidth]{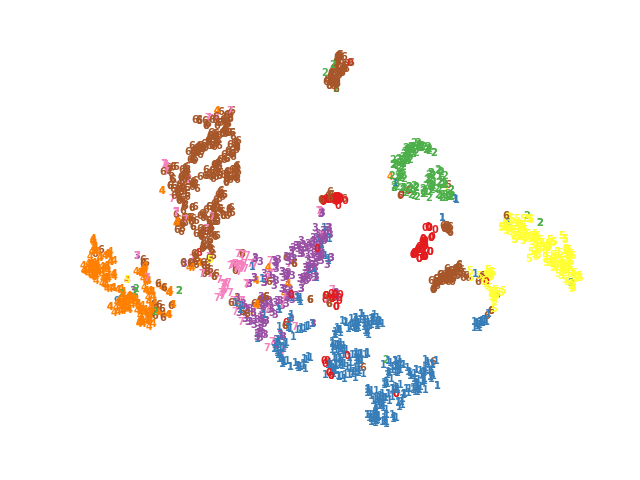}}
\vspace{3pt}
\centerline{GCA}
\end{minipage}
\begin{minipage}{0.139\linewidth}
\centerline{\includegraphics[width=\textwidth]{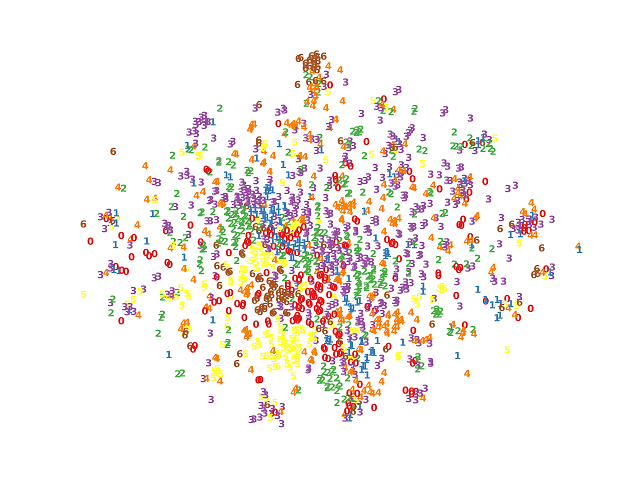}}
\vspace{3pt}
\centerline{\includegraphics[width=\textwidth]{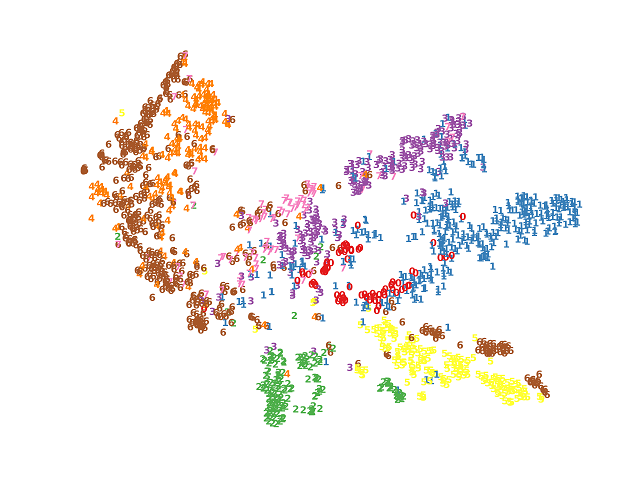}}
\vspace{3pt}
\centerline{AutoSSL}
\end{minipage}
\begin{minipage}{0.139\linewidth}
\centerline{\includegraphics[width=\textwidth]{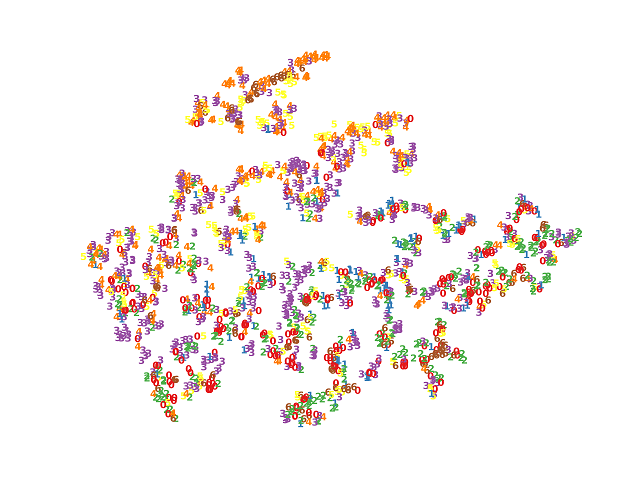}}
\vspace{3pt}
\centerline{\includegraphics[width=\textwidth]{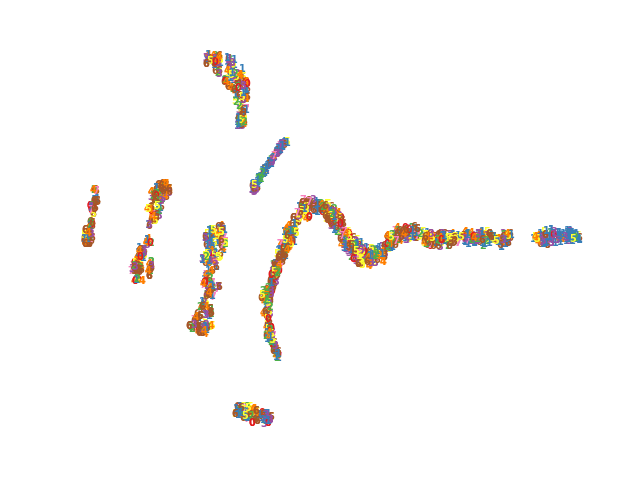}}
\vspace{3pt}
\centerline{SUBLIME}
\end{minipage}
\begin{minipage}{0.139\linewidth}
\centerline{\includegraphics[width=\textwidth]{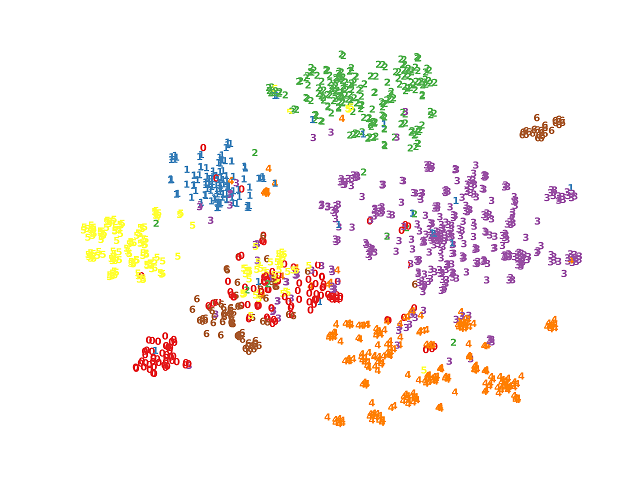}}
\vspace{3pt}
\centerline{\includegraphics[width=\textwidth]{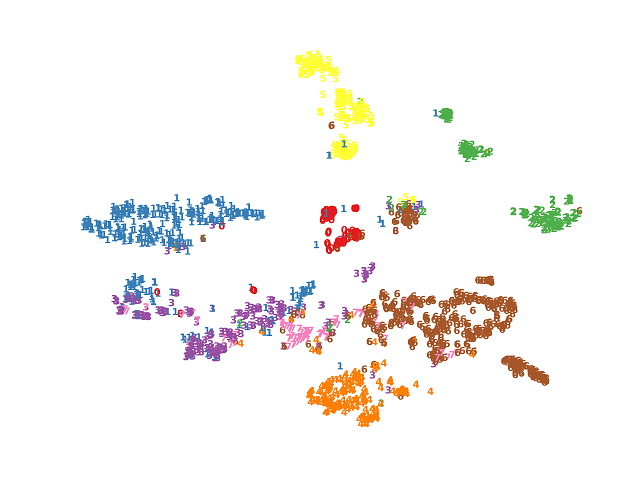}}
\vspace{3pt}
\centerline{Ours}
\end{minipage}
\caption{$T$-SNE visualization is employed to illustrate the performance of seven methods across two benchmark datasets. The first row pertains to the CORA dataset, while the second row corresponds to the AMAP dataset.}
\label{t_SNE}  
\end{figure*}

\subsection{Datasets \& Metric}

\textbf{Benchmark Datasets} We conduct extensive experiments to verify the effectiveness of CONVERT on seven benchmark datasets, including CORA \footnote{{\footnotesize{}{}\texttt{\footnotesize{}https://relational.fit.cvut.cz/dataset/CORA}{\footnotesize{}}}}, CITESEER \footnote{{\footnotesize{}{}\texttt{\footnotesize{}http://citeseerx.ist.psu.edu/index}{\footnotesize{}}}}, BAT \cite{CCGC}, EAT \cite{GCC-LDA}, UAT \cite{CCGC}, AMAP \cite{IDCRN}, and CORAFULL \cite{SCGC}. Detailed dataset statistics are summarized in Table \ref{DATASET_INFO}. 

\textbf{Evaluation Metrics} The evaluation of clustering performance encompasses four extensively utilized metrics: Accuracy (ACC), Normalized Mutual Information (NMI), Average Rand Index (ARI), and macro F1-score (F1) \cite{siwei_1,siwei_2}.

\subsection{Experimental Setup}
 
The experimental setup comprises a desktop computer equipped with an Intel Core i7-7820x CPU, an NVIDIA GeForce RTX 2080Ti GPU, 64GB RAM, and the PyTorch deep learning platform. To minimize the influence of randomness, each method is executed ten times, and the results are reported in terms of mean values with corresponding standard deviations. Training all methods is continued for 400 epochs until convergence. We employ the Adam optimizer to minimize the total loss and subsequently perform K-means on the acquired embeddings. A two-stage training strategy is adopted to ensure dependable clustering pseudo labels. In the first stage, the model's discriminative capacity is improved, while the second stage involves the utilization of high-confidence clustering pseudo labels within the matching module. Comprehensive parameter settings are elaborated in Table 1 of the Appendix.

\subsection{Performance Comparison~\textbf{(RQ1)}}

In this subsection, we conduct experiments to demonstrate the superiority of CONVERT with 12 baselines on seven datasets. To be specific, the compared clustering algorithms could roughly be divided into three classes, i.e., classical graph clustering algorithms (DAEGC \cite{DAEGC}, ARGA \cite{ARGA}, SDCN \cite{SDCN}), contrastive graph clustering algorithms (AGE \cite{AGE}, MVGRL \cite{MVGRL}, AGC-DRR \cite{AGC-DRR}), and graph augmentation clustering algorithms (GCA \cite{GCA}, AFGRL \cite{AFGRL}, AutoSSL \cite{autossl}, SUBLIME \cite{SUBLIME}, NCLA\cite{NCLA}). 

Table.\ref{compare_result} presents the attribute clustering performance comparison. From those results, we observe that 1) compared with classical deep graph clustering algorithms, our method achieves state-of-the-art performance. We conjecture the reason is that those methods rarely consider the topological information; 2) Thanks to the learnable augmentation with reliable semantic strategy, our method obtains better performance with contrastive deep graph clustering methods; 3) The graph augmentation methods achieve unpromising performance. We analyze the reason is that the semantic information of those methods drafts after the augmentations. Overall, our proposed CONVERT achieves better performance on most metrics on seven datasets. Using the EAT dataset as an example, CONVERT outperforms the second-best method by margins of 6.22\%, 5.00\%, 5.63\%, and 5.67\% in ACC, NMI, ARI, and F1 scores respectively. Furthermore, due to space constraints, more comparison results of eight baseline methods are available in Table 2 of the Appendix. These supplementary results reaffirm the efficacy of our proposed CONVERT.

\begin{table*}[]
\centering
\caption{Ablation studies of CONVERT are conducted across six datasets. The notations (w/o) $\textbf{L}\_\textbf{M}$,'' (w/o) $\textbf{R}\_\textbf{S}$,'' and ``(w/o) $\textbf{R}\_\textbf{N}$'' denote reduced models obtained by excluding the label-matching module, reliable semantic loss, and the reversible network, respectively.}
\scalebox{0.9}{
\begin{tabular}{c|c|ccc|cccc|c}
\hline
\textbf{Dataset}                   & \textbf{Metric} & \textbf{(w\/o) L\_M} & \textbf{(w\/o) R\_S} & \textbf{(w\/o) R\_N} & \textbf{Feature Mask} & \textbf{Edge Remove} & \textbf{Edge Add} & \textbf{Diffusion} & \textbf{Ours} \\ \hline
\multirow{4}{*}{\textbf{CORA}}     & \textbf{ACC}    & 73.42±1.39         & 73.13±1.42         & 73.36±1.19         & 69.56±1.91            & 65.76±3.43           & 64.77±1.95        & 70.68±1.48         & 74.07±1.51    \\
                                   & \textbf{NMI}    & 55.69±0.87         & 55.48±0.84         & 55.93±0.77         & 52.28±2.50            & 50.92±1.75           & 49.55±2.22        & 53.15±0.84         & 55.57±1.12    \\
                                   & \textbf{ARI}    & 50.35±1.38         & 49.71±1.41         & 50.24±1.22         & 45.10±2.50            & 41.46±2.21           & 40.76±1.27        & 48.39±1.34         & 50.58±2.01    \\
                                   & \textbf{F1}     & 72.62±1.93         & 72.58±1.55         & 72.75±1.30         & 68.72±2.78            & 63.94±4.30           & 62.20±2.55        & 68.66±1.67         & 72.92±3.27    \\ \hline
\multirow{4}{*}{\textbf{AMAP}}     & \textbf{ACC}    & 75.96±0.58         & 74.93±1.03         & 74.95±0.77         & 67.55±1.13            & 72.52±0.62           & 68.21±1.81        & 63.49±2.17         & 77.19±0.55    \\
                                   & \textbf{NMI}    & 65.80±1.25         & 63.76±1.76         & 65.53±1.23         & 55.66±1.32            & 59.72±0.96           & 55.25±1.97        & 51.88±1.99         & 67.20±1.07    \\
                                   & \textbf{ARI}    & 57.28±1.98         & 55.55±1.99         & 56.38±2.07         & 45.48±1.33            & 51.58±1.35           & 46.12±1.95        & 41.45±2.71         & 60.79±1.83    \\
                                   & \textbf{F1}     & 72.33±2.26         & 71.80±1.59         & 70.53±1.82         & 64.39±1.57            & 67.67±1.88           & 64.78±2.70        & 61.07±2.50         & 74.03±1.00    \\ \hline
\multirow{4}{*}{\textbf{BAT}}      & \textbf{ACC}    & 76.11±1.96         & 69.69±3.17         & 68.85±2.97         & 61.60±2.32            & 49.47±2.48           & 64.96±2.74        & 64.50±3.28         & 78.02±1.36    \\
                                   & \textbf{NMI}    & 52.04±1.86         & 45.55±2.73         & 43.96±2.82         & 36.11±2.06            & 19.44±3.09           & 39.27±3.18        & 39.84±3.61         & 53.54±1.71    \\
                                   & \textbf{ARI}    & 49.70±2.89         & 41.38±2.78         & 39.53±3.12         & 31.82±2.92            & 12.72±3.39           & 33.86±3.27        & 34.92±4.16         & 51.95±2.18    \\
                                   & \textbf{F1}     & 75.61±2.11         & 69.10±3.48         & 67.87±3.54         & 60.33±2.40            & 48.23±2.12           & 64.08±3.40        & 63.56±3.99         & 77.77±1.48    \\ \hline
\multirow{4}{*}{\textbf{EAT}}      & \textbf{ACC}    & 54.59±0.46         & 53.93±0.58         & 53.83±0.88         & 50.25±0.93            & 40.38±1.56           & 52.31±0.92        & 54.99±0.63         & 58.35±0.18    \\
                                   & \textbf{NMI}    & 28.36±0.51         & 27.49±0.70         & 27.55±0.74         & 22.24±0.99            & 12.59±1.92           & 26.29±1.30        & 27.47±0.50         & 33.36±0.16    \\
                                   & \textbf{ARI}    & 22.58±0.51         & 21.77±1.04         & 22.07±1.54         & 16.59±1.78            & 17.81±1.69           & 19.29±1.05        & 24.84±0.36         & 27.11±0.19    \\
                                   & \textbf{F1}     & 55.05±0.53         & 54.43±0.66         & 53.71±1.53         & 50.80±1.32            & 39.68±1.98           & 53.10±1.01        & 53.27±1.01         & 58.42±0.22    \\ \hline
\multirow{4}{*}{\textbf{UAT}}      & \textbf{ACC}    & 49.99±1.32         & 49.49±1.34         & 48.39±0.57         & 47.76±2.79            & 49.80±1.02           & 50.00±1.13        & 52.67±1.84         & 57.36±0.55    \\
                                   & \textbf{NMI}    & 20.93±1.67         & 20.99±1.83         & 19.93±2.39         & 21.61±2.27            & 18.81±1.03           & 24.18±1.39        & 24.29±1.65         & 28.75±1.13    \\
                                   & \textbf{ARI}    & 20.02±2.26         & 20.10±2.31         & 18.35±1.65         & 15.84±1.83            & 16.06±1.22           & 14.39±1.88        & 23.33±2.23         & 27.96±0.79    \\
                                   & \textbf{F1}     & 46.52±1.42         & 45.68±1.12         & 45.84±1.42         & 46.72±3.52            & 47.20±1.84           & 50.54±2.61        & 48.08±2.84         & 54.55±1.49    \\ \hline
\multirow{4}{*}{\textbf{CITESEER}} & \textbf{ACC}    & 64.99±1.62         & 65.00±1.62         & 64.43±1.35         & 63.62±1.10            & 66.00±1.47           & 64.16±1.06        & 65.74±0.56         & 68.43±0.69    \\
                                   & \textbf{NMI}    & 38.14±1.72         & 38.15±1.71         & 37.68±1.44         & 39.13±1.17            & 39.46±1.44           & 39.35±1.13        & 40.98±0.57         & 41.62±0.73    \\
                                   & \textbf{ARI}    & 39.12±2.32         & 39.14±2.32         & 38.20±1.78         & 37.09±1.73            & 38.66±2.24           & 37.78±1.43        & 39.66±0.91         & 42.77±1.63    \\
                                   & \textbf{F1}     & 61.22±1.38         & 61.23±1.38         & 60.82±1.62         & 60.36±0.85            & 58.50±1.24           & 60.39±1.01        & 62.00±0.81         & 62.39±2.15    \\ \hline
\end{tabular}}

\label{ablation_result}
\end{table*}
\begin{figure}[t]
\centering
\small
\begin{minipage}{0.32\linewidth}
\centerline{\includegraphics[width=1\textwidth]{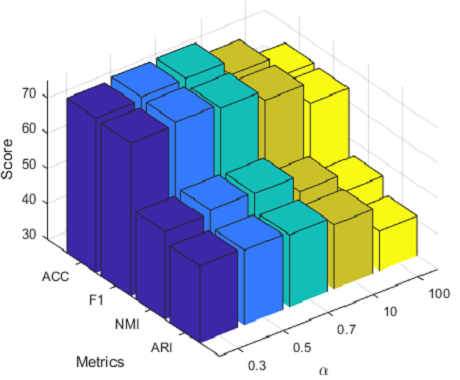}}
\vspace{3pt}
\textbf{\centerline{CORA}}
\centerline{\includegraphics[width=\textwidth]{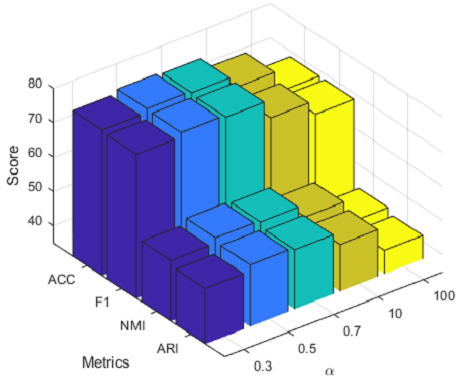}}
\vspace{3pt}
\textbf{\centerline{BAT}}
\end{minipage}
\begin{minipage}{0.32\linewidth}
\centerline{\includegraphics[width=\textwidth]{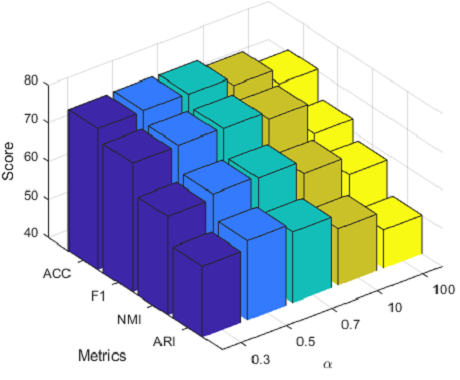}}
\vspace{3pt}
\textbf{\centerline{AMAP}}
\vspace{3pt}
\centerline{\includegraphics[width=\textwidth]{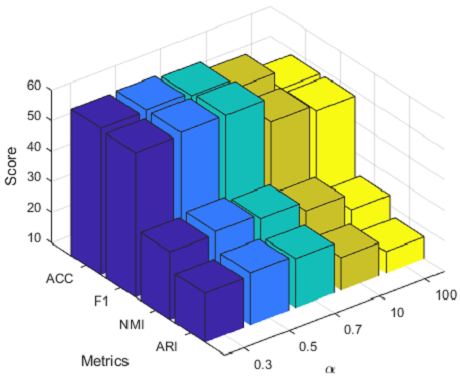}}
\vspace{3pt}
\textbf{\centerline{EAT}}
\end{minipage}
\begin{minipage}{0.32\linewidth}
\centerline{\includegraphics[width=\textwidth]{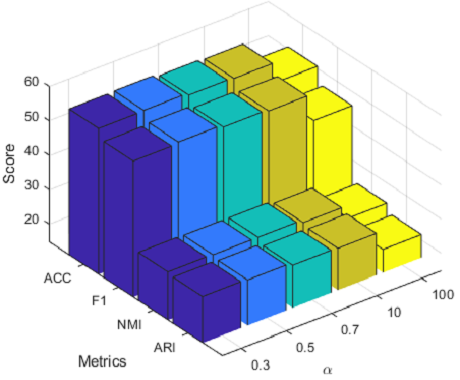}}
\vspace{3pt}
\textbf{\centerline{UAT}}
\vspace{3pt}
\centerline{\includegraphics[width=\textwidth]{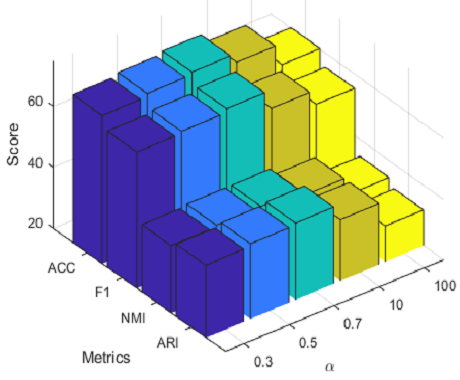}}
\vspace{3pt}
\textbf{\centerline{CITESEER}}
\end{minipage}
\caption{Sensitivity analysis of the hyper-parameter $\alpha$.}
\label{sen_alpha}
\end{figure}
\begin{figure}[t]
\centering
\small
\begin{minipage}{0.32\linewidth}
\centerline{\includegraphics[width=1\textwidth]{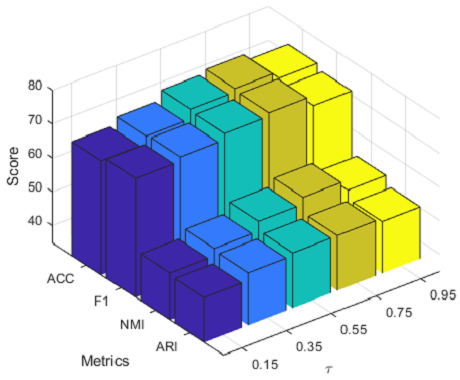}}
\vspace{3pt}
\textbf{\centerline{CORA}}
\centerline{\includegraphics[width=\textwidth]{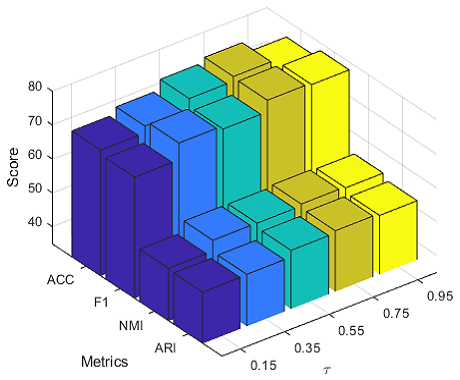}}
\vspace{3pt}
\textbf{\centerline{BAT}}
\end{minipage}
\begin{minipage}{0.32\linewidth}
\centerline{\includegraphics[width=\textwidth]{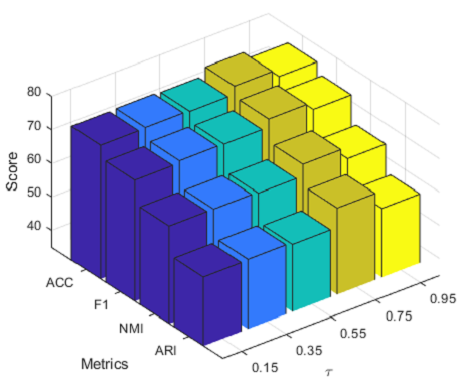}}
\vspace{3pt}
\textbf{\centerline{AMAP}}
\vspace{3pt}
\centerline{\includegraphics[width=\textwidth]{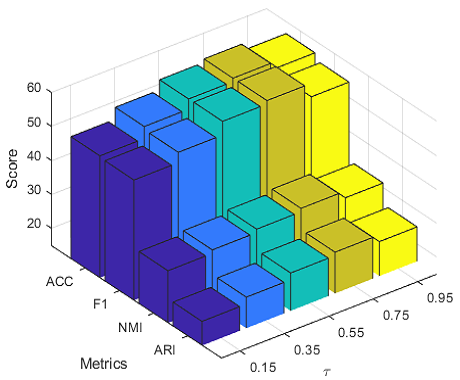}}
\vspace{3pt}
\textbf{\centerline{EAT}}
\end{minipage}
\begin{minipage}{0.32\linewidth}
\centerline{\includegraphics[width=\textwidth]{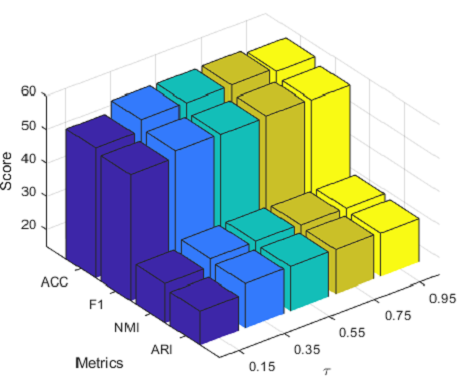}}
\vspace{3pt}
\textbf{\centerline{UAT}}
\vspace{3pt}
\centerline{\includegraphics[width=\textwidth]{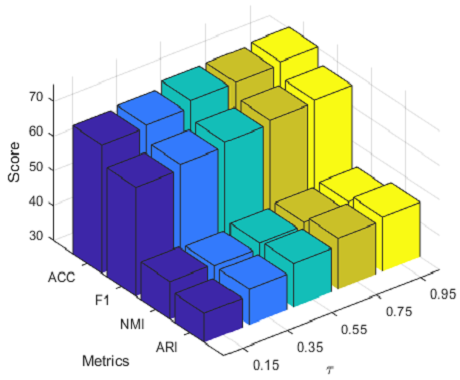}}
\vspace{3pt}
\textbf{\centerline{CITESEER}}
\end{minipage}
\caption{Sensitivity analysis of the hyper-parameter $\tau$.}
\label{sen_tau}
\end{figure}

\subsection{Ablation Studies~\textbf{(RQ2)}}

\subsubsection{Effectiveness of the proposed module}

As shown in Table.\ref{ablation_result}, we conduct experiments to verify the effectiveness of the reversible network. To be specific, we adopt ``(w/o) $\textbf{L}\_\textbf{M}$'', ``(w/o) $\textbf{R}\_\textbf{S}$'' and ``(w/o) $\textbf{R}\_\textbf{N}$'' to denote the reduced models by removing the label matching module, the reliable semantic loss, the reversible network respectively. Without any of our proposed modules, the clustering performance will decrease, indicating that each module makes contributions to boosting the performance. We further analyze the reasons are as follows: 1) the network is better guided by the high-confidence clustering pseudo labels through the label-matching mechanism. 2) The proposed semantic loss and the reversible network guarantee the reliability of semantic information, thus avoiding semantic drifting.

\begin{table*}[]
\centering
\caption{Training Time Comparison on six datasets. Avg. denotes the average time cost on six datasets. Besides, OOM represents Out-Of-Memory during the training process.}
\scalebox{0.8}{
\begin{tabular}{c|ccccccccc|c}
\hline
\multirow{2}{*}{\textbf{Dataset}} & \textbf{DEC}       & \textbf{AE}        & \textbf{DAEGC}      & \textbf{MGAE}        & \textbf{AGE}         & \textbf{SDCN}     & \textbf{MVGRL}     & \textbf{MCGC}         & \textbf{SCAGC}    & \textbf{CONVERT} \\ \cline{2-11} 
                                  & \textbf{ICML 2016} & \textbf{ICML 2017} & \textbf{IJCAI 2019} & \textbf{SIGKDD 2019} & \textbf{SIGKDD 2020} & \textbf{WWW 2020} & \textbf{ICML 2020} & \textbf{NeurIPS 2021} & \textbf{TMM 2022} & \textbf{Ours}  \\ \hline
\textbf{CORA}                     & 91.13              & 47.31              & 12.97               & 7.38                 & 46.65                & 11.32             & 14.72              & 118.07                & 54.08             & 16.89          \\
\textbf{CITESEER}                 & 223.95             & 74.69              & 14.70               & 6.69                 & 70.63                & 11.00             & 18.31              & 126.06                & 50.00             & 57.23          \\
\textbf{BAT}                      & 21.37              & 7.46               & 4.79                & 3.83                 & 2.49                 & 11.50             & 3.19               & 2.28                  & 93.79             & 5.15           \\
\textbf{EAT}                      & 26.99              & 9.56               & 5.14                & 4.64                 & 3.86                 & 12.12             & 3.32               & 2.87                  & 47.79             & 5.18           \\
\textbf{UAT}                      & 42.30              & 29.57              & 6.44                & 4.75                 & 8.95                 & 10.64             & 4.27               & 23.10                 & 64.70             & 6.26           \\
\textbf{AMAP}                     & 264.20             & 94.48              & 39.62               & 18.64                & 377.49               & 19.28             & 131.38             & OOM                   & 150.54            & 46.98          \\ \hline
\textbf{Avg.}                     & 111.66             & 43.85              & 13.94               & 7.66                 & 85.01                & 12.64             & 29.20              & -                     & 76.82             & 22.95          \\ \hline
\end{tabular}}
\label{time_compare}
\end{table*}

\subsubsection{Effectiveness of the Learnable Augmentation Module}
To validate the effectiveness of the learnable augmentation module, we conduct the ablation studies shown in Table.\ref{ablation_result}. Here, we adopt the same backbone for all experiments and four commonly used graph augmentations, e.g., randomly masking 10\% feature (Feature Mask), randomly dropping 10\% graph edges (Edge Remove), randomly adding 10\% graph edges (Edge Add), and graph diffusion with 0.10 teleportation rate (Diffusion). From the results, we conclude that since utilizing the common graph augmentations, the clustering performance is limited by the drifting semantic \cite{AFGRL}. In summary, the experiment results demonstrate the effectiveness of the learnable augmentation module.

\begin{figure}
\centering
\scalebox{0.35}{
\includegraphics{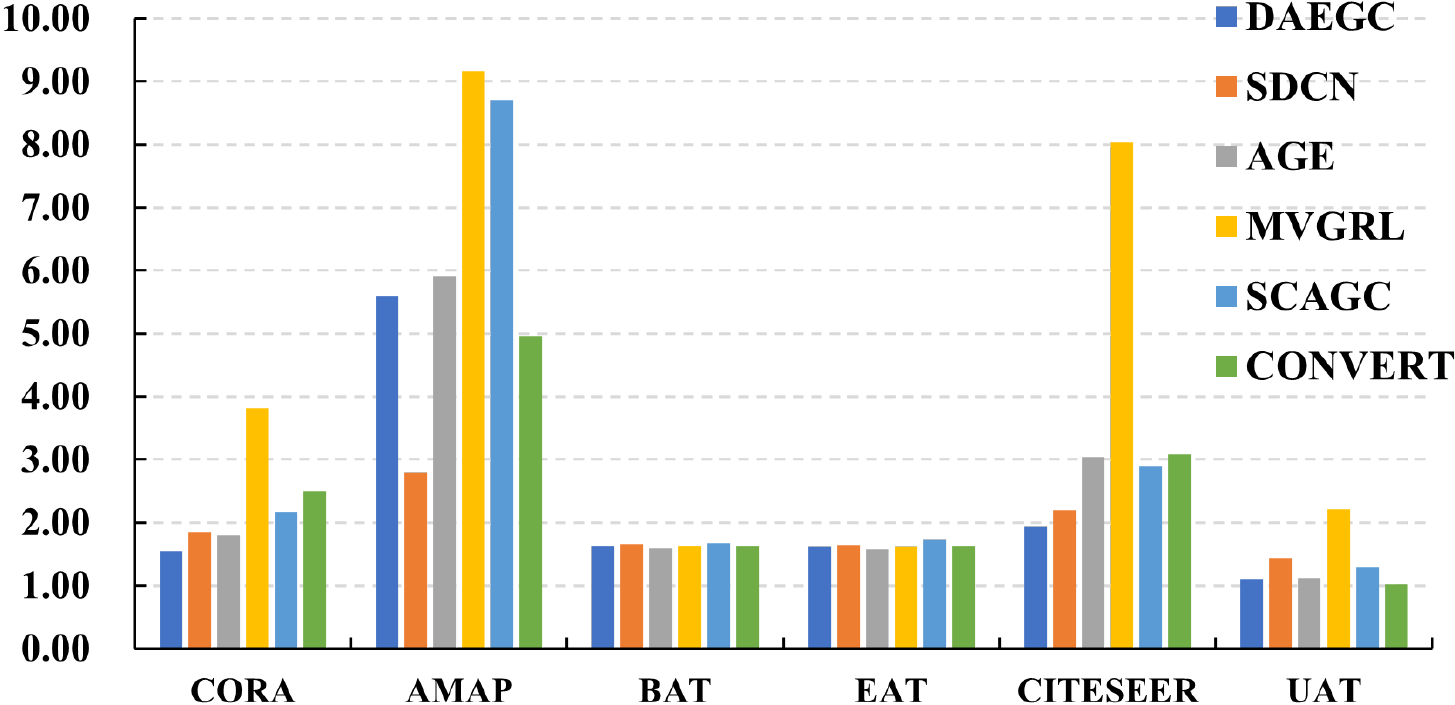}}
\caption{Illustration of the gpu memory cost of CONVERT with five algorithms on six datasets.}
\label{gpu_cost}
\end{figure}

\subsection{Efficiency analysis~\textbf{(RQ3)}}

In this subsection, we compare the efficiency of CONVERT and other state-of-the-art clustering algorithms. As can be observed in Fig.\ref{gpu_cost} and Table.\ref{time_compare}, CONVERT has a comparable GPU memory cost and training time with other clustering methods. In summary, the efficiency of CONVERT is acceptable. We analyze the reason is that we adopt the graph filter to extract the feature, avoiding the complex convolution and aggregation operations.

\subsection{Hyper-parameter Analysis~\textbf{(RQ4)}}

\subsubsection{Sensitivity Analysis of hyper-parameter $\alpha$}

As can be observed in Fig.\ref{sen_alpha}, we observe that the performance of CONVERT will not fluctuate greatly when the $\alpha \in [0.3,0.7]$. When the value of $\alpha$ drastically changes, the balance of the model will be destroyed, thus limiting the clustering performance. Moreover, we investigate the influence of $\beta$. CONVERT is not sensitive to $\beta$. Experiments can be found in Fig. 1 in Appendix.


\subsubsection{Sensitivity Analysis of hyper-parameter $\tau$}

Besides, we investigate the impact of $\tau$. As shown in Fig.\ref{sen_tau}, we observe that CONVERT achieve better performance when $\tau \in [0.55, 0.75]$. There are two reasons. Firstly, the discriminative capacity of CONVERT is limited when $\tau < 0.55$ since the low confidence of the pseudo labels. Secondly, when $\tau > 0.75$, the model easily leads to confirmation bias due to the over-confidence pseudo labels \cite{confirmation}.

\subsection{Visualization Analysis~\textbf{(RQ5)}}

To unveil the inherent clustering structure, this subsection employs visualization to depict the distribution of the learned embeddings. Specifically, experiments are conducted using the $t$-SNE algorithm \cite{T_SNE} on CORA and AMAP datasets. As depicted in Figure \ref{t_SNE}, the visual results highlight that CONVERT exhibits an enhanced clustering structure.

\section{Conclusion}

In this paper, we present a learnable augmentation strategy for attribute clustering with reliable semantics, termed CONVERT. Specifically, we design a reversible network to generate augmented views. The perturb-recover embedding operation avoids semantic drift. Then, we design a semantic loss to further guarantee the reliability of the semantic. Moreover, we propose a label-matching mechanism to align the semantic labels and high-confidence pseudo labels, thus utilizing the clustering information to guide the model. Extensive experiments have demonstrated the effectiveness of CONVERT.

\begin{acks}
This work was supported by the National Key R$\&$D Program of China 2020AAA0107100 and the National Natural Science Foundation of China (project no. 62325604, 62276271). 
\end{acks}

\bibliographystyle{ACM-Reference-Format}
\bibliography{ref}
\end{document}


\appendix
\title{Appendix of CONVERT:~Contrastive Graph Clustering with Reliable Augmentation}
\ccsdesc[500]{Computing methodologies~Cluster analysis}
\keywords{Attribute Graph Clustering, Contrastive Learning}
\maketitle




\section{Details of the Proposed Method}

We introduce the implementation of our proposed method CONVERT with PyTorch-style pseudo codes in Algorithm \ref{code}. 

\begin{algorithm}[h]
	\caption{PyTorch-style Pseudo Code of Our Method.}
	\label{code}
	\definecolor{codeblue}{rgb}{0.25,0.5,0.5}
	\lstset{
		backgroundcolor=\color{white},
		basicstyle=\fontsize{7.2pt}{7.2pt}\ttfamily\selectfont,
		columns=fullflexible,
		breaklines=true,
		captionpos=b,
		commentstyle=\fontsize{7.2pt}{7.2pt}\color{codeblue},
		keywordstyle=\fontsize{7.2pt}{7.2pt},
	}
    
    \begin{lstlisting}[language=python]
    # X: Original Attribute
    # A: Original Structure
    # p: Perturb Network
    # r: Recover Network
    # h: High-confidence Pseudo Labels
    # N: High-confidece Epoch
    # CE: Cross-Entropy Function
    # L_c: NT-Xent loss 
    # L_s: Semantic loss
    # alpha, beta: trade-off parameters
    # clu_res: Cluster Results
    for epoch in range(epoch_num):

        #Smooth the feature
        X1 = graph_filter(X, A)

        # Perturb the Attribute Matrix
        X2 = p(X1)
        
        # Net Encoding
        E1 = F.normalization(X1,dim=1,p=2)
        E2 = F.normalization(X2,dim=1,p=2)

        #Perturb and Recover Embeddings
        H1 = r(E2)
        H2 = p(E1)

        # Calculate the similarity of embeddings 
        S_r = E1 * H1
        S_p = E2 * H2
        
        L_s = MSE(S_r, S_p)
        
        # Clustering and High-confidence Pseudo Label
        clu_res, h = clustering((E1+E2/2))
        
        loss_c = L_c (E1, H1) + L_c (E2, H2)
        loss = loss_c + alpha * L_s
        
        if epoch > N:
            # Label Matching
            L_m = CE(h, softmax(E1)) + CE(h, softmax(E2))
            loss = loss_c + alpha * L_s + beta * L_m
        # optimization
        loss.backward()
        optimizer.step()
    clu_res, _, _ = clustering((E1+E2/2))
    return clu_res
    
    
    
    \end{lstlisting}
    
\end{algorithm}

\begin{figure}[h]
\centering
\small
\begin{minipage}{0.32\linewidth}
\centerline{\includegraphics[width=1\textwidth]{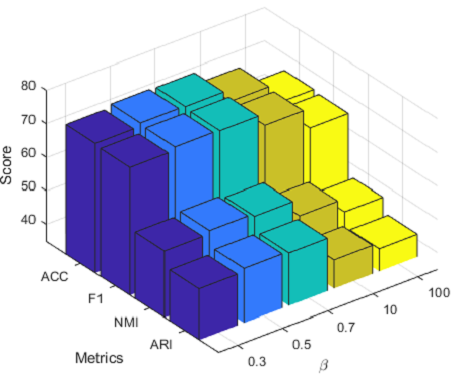}}
\vspace{3pt}
\textbf{\centerline{CORA}}
\centerline{\includegraphics[width=\textwidth]{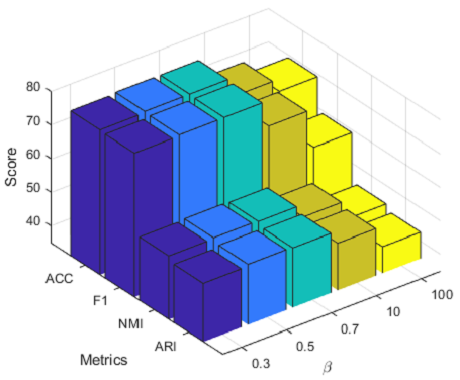}}
\vspace{3pt}
\textbf{\centerline{BAT}}
\end{minipage}
\begin{minipage}{0.32\linewidth}
\centerline{\includegraphics[width=\textwidth]{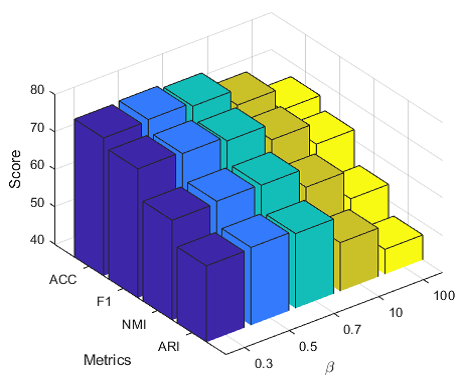}}
\vspace{3pt}
\textbf{\centerline{AMAP}}
\vspace{3pt}
\centerline{\includegraphics[width=\textwidth]{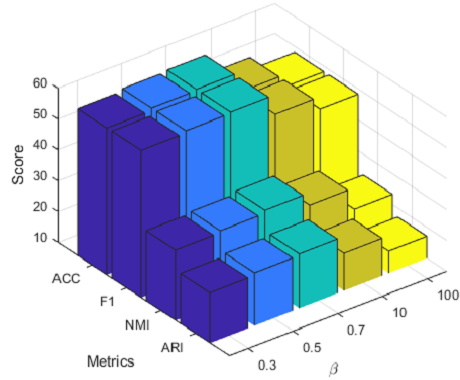}}
\vspace{3pt}
\textbf{\centerline{EAT}}
\end{minipage}
\begin{minipage}{0.32\linewidth}
\centerline{\includegraphics[width=\textwidth]{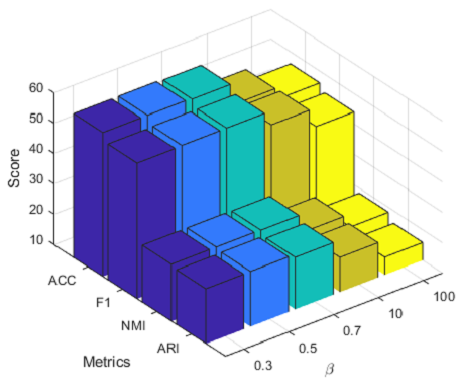}}
\vspace{3pt}
\textbf{\centerline{UAT}}
\vspace{3pt}
\centerline{\includegraphics[width=\textwidth]{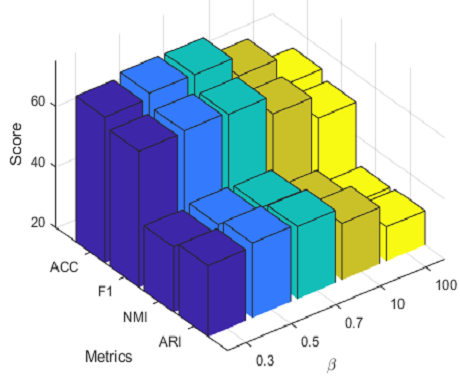}}
\vspace{3pt}
\textbf{\centerline{CITESEER}}
\end{minipage}
\caption{Sensitivity analysis of the hyper-parameter $\beta$.}
\label{sen_beta}
\end{figure}

\begin{table*}[!t]
\centering
\caption{Statistics and hyper-parameter settings of six benchmark datasets.}
\small
\scalebox{1.}{
\begin{tabular}{c|cclclclclclclcl}
\hline
                                           & \textbf{Dataset}       & \multicolumn{2}{c}{\textbf{CORA}} & \multicolumn{2}{c}{\textbf{CITE}} & \multicolumn{2}{c}{\textbf{AMAP}} & \multicolumn{2}{c}{\textbf{BAT}} & \multicolumn{2}{c}{\textbf{EAT}} & \multicolumn{2}{c}{\textbf{UAT}}  & \multicolumn{2}{c}{\textbf{Corafull}} \\ \hline
\multirow{5}{*}{\textbf{Statistics}}       & Type                   & \multicolumn{2}{c}{Graph}         & \multicolumn{2}{c}{Graph}         & \multicolumn{2}{c}{Graph}         & \multicolumn{2}{c}{Graph}        & \multicolumn{2}{c}{Graph}        & \multicolumn{2}{c}{Graph} & \multicolumn{2}{c}{Graph}       \\
                                           & \# Samples             & \multicolumn{2}{c}{2708}          & \multicolumn{2}{c}{3327}          & \multicolumn{2}{c}{7650}          & \multicolumn{2}{c}{131}          & \multicolumn{2}{c}{399}          & \multicolumn{2}{c}{1190}   & \multicolumn{2}{c}{19793}      \\
                                           & \# Dimensions          & \multicolumn{2}{c}{1433}          & \multicolumn{2}{c}{3703}          & \multicolumn{2}{c}{745}           & \multicolumn{2}{c}{81}           & \multicolumn{2}{c}{203}          & \multicolumn{2}{c}{239} & \multicolumn{2}{c}{8710}         \\
                                           & \# Edges               & \multicolumn{2}{c}{5429}          & \multicolumn{2}{c}{4732}          & \multicolumn{2}{c}{119081}        & \multicolumn{2}{c}{1038}         & \multicolumn{2}{c}{5994}         & \multicolumn{2}{c}{13599}  & \multicolumn{2}{c}{63421}        \\
                                           & \# Classes             & \multicolumn{2}{c}{7}             & \multicolumn{2}{c}{6}             & \multicolumn{2}{c}{8}             & \multicolumn{2}{c}{4}            & \multicolumn{2}{c}{4}            & \multicolumn{2}{c}{4}   & \multicolumn{2}{c}{70}           \\ \hline
\multirow{4}{*}{\makecell[c]{\textbf{Hyper-} \\ \textbf{parameters}}} & $\tau$          & \multicolumn{2}{c}{0.75}           & \multicolumn{2}{c}{0.75}           & \multicolumn{2}{c}{0.75}           & \multicolumn{2}{c}{0.75}          & \multicolumn{2}{c}{0.75}          & \multicolumn{2}{c}{0.75}   & \multicolumn{2}{c}{0.75}        \\
                                           & $\alpha$                   & \multicolumn{2}{c}{0.5}             & \multicolumn{2}{c}{0.5}             & \multicolumn{2}{c}{0.5}             & \multicolumn{2}{c}{0.5}            & \multicolumn{2}{c}{0.5}            & \multicolumn{2}{c}{0.5}   & \multicolumn{2}{c}{0.5}          \\
                                           & $\beta$                      & \multicolumn{2}{c}{0.5}             & \multicolumn{2}{c}{0.5}             & \multicolumn{2}{c}{0.5}             & \multicolumn{2}{c}{0.5}            & \multicolumn{2}{c}{0.5}            & \multicolumn{2}{c}{0.5} & \multicolumn{2}{c}{0.5}            \\
                                           & Learning rate & \multicolumn{2}{c}{$10^{-5}$}            & \multicolumn{2}{c}{$10^{-3}$}            & \multicolumn{2}{c}{$10^{-3}$}            & \multicolumn{2}{c}{$10^{-3}$}           & \multicolumn{2}{c}{$10^{-7}$}           & \multicolumn{2}{c}{$10^{-3}$}    & \multicolumn{2}{c}{$10^{-4}$}        \\ \hline
\end{tabular}}
\label{DATASET_INFO}
\end{table*}

\begin{table*}[]
\centering
\caption{Additional comparison experiments on six benchmark datasets. The clustering performance is evaluated by four metrics with mean value and standard deviation.}
\scalebox{0.7}{
\begin{tabular}{c|c|cccc|ccccc}
\hline
                                    &                                   & \multicolumn{4}{c|}{\textbf{Deep   Clustering}}                                                  & \multicolumn{5}{c}{\textbf{Deep   Graph Clustering}}                                                                        \\ \cline{3-11} 
                                    &                                   & \textbf{MGAE}      & \textbf{DCN}       & \textbf{DEC}       & \textbf{AdaGAE}                   & \textbf{DFCN}      & \textbf{GDCL}       & \textbf{SLAPS}     & \multicolumn{1}{c|}{\textbf{DCRN}}         & \textbf{CONVERT} \\
\multirow{-3}{*}{\textbf{Dataset}}  & \multirow{-3}{*}{\textbf{Metric}} & \textbf{CIKM 2019} & \textbf{ICML 2017} & \textbf{ICML 2016} & \textbf{TPAMI 2021}               & \textbf{AAAI 2021} & \textbf{IJCAI 2021} & \textbf{NIPS 2021} & \multicolumn{1}{c|}{\textbf{AAAI 2022}}    & \textbf{Ours}  \\ \hline
                                    & \textbf{ACC}                      & 43.38±2.11         & 49.38±0.91         & 46.50±0.26         & 50.06±1.58                        & 36.33±0.49         & 70.83±0.47          & 64.21±0.12         & \multicolumn{1}{c|}{61.93±0.47}            & 74.07±1.51     \\
                                    & \textbf{NMI}                      & 28.78±2.97         & 25.65±0.65         & 23.54±0.34         & 32.19±1.34                        & 19.36±0.87         & 56.60±0.36          & 41.16±1.24         & \multicolumn{1}{c|}{45.13±1.57}            & 55.57±1.12     \\
                                    & \textbf{ARI}                      & 16.43±1.65         & 21.63±0.58         & 15.13±0.42         & 28.25±0.98                        & 04.67±2.10         & 48.05±0.72          & 35.96±0.65         & \multicolumn{1}{c|}{33.15±0.14}            & 50.58±2.01     \\
\multirow{-4}{*}{\textbf{CORA}}     & \textbf{F1}                       & 33.48±3.05         & 43.71±1.05         & 39.23±0.17         & 53.53±1.24                        & 26.16±0.50         & 52.88±0.97          & 63.72±0.26         & \multicolumn{1}{c|}{49.50±0.42}            & 72.92±3.27     \\ \hline
                                    & \textbf{ACC}                      & 71.57±2.48         & 48.25±0.08         & 47.22±0.08         & 67.70±0.54                        & 76.82±0.23         & 43.75±0.78          & 60.09±1.14         & \multicolumn{1}{c|}{}                      & 77.19±0.55     \\
                                    & \textbf{NMI}                      & 62.13±2.79         & 38.76±0.30         & 37.35±0.05         & 55.96±0.87                        & 66.23±1.21         & 37.32±0.28          & 51.15±0.87         & \multicolumn{1}{c|}{}                      & 67.20±1.07     \\
                                    & \textbf{ARI}                      & 48.82±4.57         & 20.80±0.47         & 18.59±0.04         & 46.20±0.45                        & 58.28±0.74         & 21.57±0.51          & 42.87±0.75         & \multicolumn{1}{c|}{}                      & 60.79±1.83     \\
\multirow{-4}{*}{\textbf{AMAP}}     & \textbf{F1}                       & 68.08±1.76         & 47.87±0.20         & 46.71±0.12         & 62.95±0.74                        & 71.25±0.31         & 38.37±0.29          & 47.73±0.98         & \multicolumn{1}{c|}{\multirow{-4}{*}{OOM}} & 74.03±1.00     \\ \hline
                                    & \textbf{ACC}                      & 53.59±2.04         & 47.79±3.95         & 42.09±2.21         & 43.51±0.48                        & 55.73±0.06         & 45.42±0.54          & 41.22±1.25         & \multicolumn{1}{c|}{67.94±1.45}            & 78.02±1.36     \\
                                    & \textbf{NMI}                      & 30.59±2.06         & 18.03±7.73         & 14.10±1.99         & 15.84±0.78                        & 48.77±0.51         & 31.70±0.42          & 17.05±0.87         & \multicolumn{1}{c|}{47.23±0.74}            & 53.54±1.71     \\
                                    & \textbf{ARI}                      & 24.15±1.70         & 13.75±6.05         & 07.99±1.21         & 07.80±0.41                        & 37.76±0.23         & 19.33±0.57          & 06.86±2.14         & \multicolumn{1}{c|}{39.76±0.87}            & 51.95±2.18     \\
\multirow{-4}{*}{\textbf{BAT}}      & \textbf{F1}                       & 50.83±3.23         & 46.80±3.44         & 42.63±2.35         & 43.15±0.77                        & 50.90±0.12         & 39.94±0.57          & 37.64±0.57         & \multicolumn{1}{c|}{67.40±0.35}            & 77.77±1.48     \\ \hline
                                    & \textbf{ACC}                      & 44.61±2.10         & 38.85±2.32         & 36.47±1.60         & 32.83±1.24                        & 49.37±0.19         & 33.46±0.18          & 48.62±1.65         & \multicolumn{1}{c|}{50.88±0.55}            & 58.35±0.18     \\
                                    & \textbf{NMI}                      & 15.60±2.30         & 06.92±2.80         & 04.96±1.74         & 04.36±1.87                        & 32.90±0.41         & 13.22±0.33          & 28.33±2.56         & \multicolumn{1}{c|}{22.01±1.23}            & 33.36±0.16     \\
                                    & \textbf{ARI}                      & 13.40±1.26         & 05.11±2.65         & 03.60±1.87         & 02.47±0.54                        & 23.25±0.18         & 04.31±0.29          & 24.59±0.58         & \multicolumn{1}{c|}{18.13±0.85}            & 27.11±0.19     \\
\multirow{-4}{*}{\textbf{EAT}}      & \textbf{F1}                       & 43.08±3.26         & 38.75±2.25         & 34.84±1.28         & 32.39±0.47                        & 42.95±0.04         & 25.02±0.21          & 40.42±1.44         & \multicolumn{1}{c|}{47.06±0.66}            & 58.42±0.22     \\ \hline
                                    & \textbf{ACC}                      & 48.97±1.52         & 46.82±1.14         & 45.61±1.84         & 52.10±0.87                        & 33.61±0.09         & 48.70±0.06          & 49.77±1.24         & \multicolumn{1}{c|}{48.70±0.06}            & 57.36±0.55     \\
                                    & \textbf{NMI}                      & 20.69±0.98         & 17.18±1.60         & 16.63±2.39         & 26.02±0.71                        & 26.49±0.41         & 25.10±0.01          & 12.86±0.65         & \multicolumn{1}{c|}{25.10±0.01}            & 28.75±1.13     \\
                                    & \textbf{ARI}                      & 18.33±1.79         & 13.59±2.02         & 13.14±1.97         & {\color[HTML]{000000} 24.47±0.13} & 11.87±0.23         & 21.76±0.01          & 17.36±0.98         & \multicolumn{1}{c|}{21.76±0.01}            & 27.96±0.79     \\
\multirow{-4}{*}{\textbf{UAT}}      & \textbf{F1}                       & 47.95±1.52         & 45.66±1.49         & 44.22±1.51         & 43.44±0.85                        & 25.79±0.29         & 45.69±0.08          & 10.56±1.34         & \multicolumn{1}{c|}{45.69±0.08}            & 54.55±1.49     \\ \hline
                                    & \textbf{ACC}                      & 61.35±0.80         & 57.08±0.13         & 55.89±0.20         & 54.01±1.11                        & 69.50±0.20         & 66.39±0.65          & 64.76±0.07         & \multicolumn{1}{c|}{66.39±0.65}            & 68.43±0.69     \\
                                    & \textbf{NMI}                      & 34.63±0.65         & 27.64±0.08         & 28.34±0.30         & 27.79±0.47                        & 43.90±0.20         & 39.52±0.38          & 39.11±0.06         & \multicolumn{1}{c|}{39.52±0.38}            & 41.62±0.73     \\
                                    & \textbf{ARI}                      & 33.55±1.18         & 29.31±0.14         & 28.12±0.36         & 24.19±0.85                        & 45.50±0.30         & 41.07±0.96          & 37.54±0.12         & \multicolumn{1}{c|}{41.07±0.96}            & 42.77±1.63     \\
\multirow{-4}{*}{\textbf{CITESEER}} & \textbf{F1}                       & 57.36±0.82         & 53.80±0.11         & 52.62±0.17         & 51.11±0.64                        & 64.30±0.20         & 61.12±0.70          & 59.64±0.05         & \multicolumn{1}{c|}{61.12±0.70}            & 62.39±2.15     \\ \hline
\end{tabular}}
\label{compare}
\end{table*}

\section{Additional Experiments}

Due to the limitation of the original paper pages, in this section, we have conducted additional experiments to further the superiority of our proposed CONVERT, i.e., Sensitivity analysis about $\beta$, comparison experiments with other clustering methods.

We have conducted additional experiments to further the effectiveness of the proposed CONVERT. Due to the limitation of the original paper pages, in this section, we conduct additional experiments including comparison experiments and visualization analysis experiments.

\subsection{Sensitivity Analysis of hyper-parameter $\beta$}

As can be observed in Fig.\ref{sen_beta}, we observe that the performance of CONVERT will not fluctuate greatly when the $\alpha \in [0.3,0.7]$. When the value of $\beta$ drastically change, the balance of the model will be destroyed, thus limiting the clustering performance. 

\subsection{Statistics and Hyper-parameter Settings}
To guarantee reproducibility, we report the statistics summary and hyper-parameter settings of our proposed method in Table \ref{DATASET_INFO}.

\subsection{Additional Comparison Experiments}

Due to the limitation of the pages,in this section, we have conducted additional experiments to further the superiority of our proposed CONVERT. Specifically, two categories methods are compared in this section, i.e. deep clustering methods (MGAE \cite{MGAE}, DCN \cite{AE_K_MEANS}, DEC \cite{DEC}, AdaGAE \cite{AdaGAE}), and deep graph clustering methods (DFCN \cite{DFCN}, GDCL \cite{GDCL}, SLAPS \cite{SLAPS}, DCRN \cite{DCRN}). The experiment results are shown in Table.\ref{compare}. We could observe as follows. 1) Deep clustering methods are not comparable with our proposed methods. We conjecture that those methods overlook the graph structure. 2) CONVERT could achieve better performance than deep graph clustering methods. The reason is that the proposed learnable augmentation strategy ensures the semantic reliability of the augmented view, thus enhancing the supervision information capture capability of our method.

\bibliographystyle{ACM-Reference-Format}
\bibliography{ref}